\documentclass[letterpaper,twocolumn,10pt]{article}
\usepackage{usenix2019_v3}

\usepackage{tikz}
\usepackage{amsmath}
\DeclareMathOperator*{\argmax}{arg\,max}
\DeclareMathOperator*{\argmin}{arg\,min}

\usepackage{booktabs}       
\usepackage{amsfonts}       
\usepackage{microtype}
\usepackage{booktabs}       

\usepackage{subfig}
\usepackage{wrapfig}
\usepackage{booktabs, multirow} 

\usepackage{filecontents}

\usepackage{bbding}
\usepackage{pifont}
\usepackage{wasysym}
\usepackage{amssymb}

\usepackage{algorithmic}
\usepackage[ruled]{algorithm2e}

\usepackage{authblk}

\definecolor{ao(english)}{rgb}{0.0, 0.5, 0.0}
\definecolor{applegreen}{rgb}{0.55, 0.71, 0.0}
\def\Equal{\texttt{=}}

\newcommand*\circled[1]{\tikz[baseline=(char.base)]{
            \node[shape=circle,draw,inner sep=1pt] (char) {\bfseries\footnotesize #1};}}

\newcommand{\bcircled}[1]{\tikz[baseline=(myanchor.base)] \node[circle,fill=.,inner sep=1pt] (myanchor) {\color{-.}\bfseries\footnotesize #1};}

\begin{document}

\date{}

\title{\Large \bf Learning to Separate Clusters of Adversarial Representations \\ for Robust Adversarial Detection}

\author[1]{Byunggill Joe} 
\author[2]{Jihun Hamm} 
\author[1]{Sung Ju Hwang} 
\author[1]{Sooel Son} 
\author[1]{Insik Shin} 
\affil[1]{\it{School of Computing, KAIST}}
\affil[2]{\it{Tulane University}}
\affil[ ]{\{cp4419,sjhwang82,sl.son,insik.shin\}@kaist.ac.kr, jhamm3@tulane.edu}

\maketitle
\begin{abstract}
    Deep neural networks have shown promising performances on various tasks, but they are also susceptible to incorrect predictions induced by imperceptibly small perturbations in inputs. Although a large number of previous works have 
been proposed to detect adversarial examples, most of them cannot effectively detect 
adversarial examples against adaptive whitebox attacks 
where an adversary has the knowledge of the victim model
and the defense method. In this paper, we propose a new probabilistic adversarial detector which is effective against the adaptive whitebox attacks.
From the observation that the distributions of adversarial and benign examples are highly overlapping in the output space of the last layer of the neural network, we hypothesize that this overlapping representation limits the effectiveness of existing adversarial detectors fundamentally, and we aim to reduce the overlap by separating adversarial and benign representations.
To realize this, we propose a training algorithm 
to maximize the likelihood of adversarial representations
using a deep encoder and a Gaussian mixture model.
We thoroughly validate the robustness of our detectors 
with 7 different adaptive whitebox and blackbox attacks
and 4 perturbation types $\ell_p, p \in \{0, 1, 2, \infty\}$.
The results show our detector improves the worst case 
Attack Success Ratios up to 45\% (MNIST) and 27\% (CIFAR10) compared to existing detectors.
\end{abstract}

\section{Introduction}
In the past decade, deep neural networks have achieved impressive performance on many tasks including image classification. Recently, it was discovered that imperceptibly small perturbations called adversarial examples can reduce the accuracy of classification models to almost zero without inducing semantic changes in the original images~\cite{szegedy,fgsm}.
A number of defense methods have been proposed to overcome the attack of adversarial examples, including the ones of improving the robustness of inference under input perturbation~\cite{dae,stability,pgd,defensegan,sap,towardmnist} and detecting  adversarial vs clean examples~\cite{kd-detect,gong-detect,grosse-detect,rce,feature-squeeze,mutation-detect,nic}. Despite the effectiveness of those methods under blackbox attack scenarios, many of them are ineffective under adaptive whitebox scenarios in which the adversary can create sophisticated adversarial examples using the full knowledge of the defender~\cite{easily,obfuscated}.


In this paper, we propose a robust adversarial detector 
under the adaptive whitebox threat model, providing a new perspective to examine and address adversarial attacks in the \textit{representation} space -- the output from the last layer in a neural network (before softmax normalization).
In our preliminary experiment (Section \ref{motivation_section}),
we find the distribution of adversarial examples
overlaps the distribution of benign examples in the representation space.
The overlap between benign and adversarial distributions makes it hardly possible to distinguish adversarial examples through distance-based detection methods in the representation space.
This observation motivates us to explore the possibility to separate benign and adversarial distributions in the representation space. This presents a novel perspective on adversarial examples different from previous perspectives which view them simply as random examples that fall onto the incorrect side of the decision boundary of classifiers ~\cite{gong-detect,grosse-detect}.

\begin{figure*}
\centering

\includegraphics[width=15.5cm]{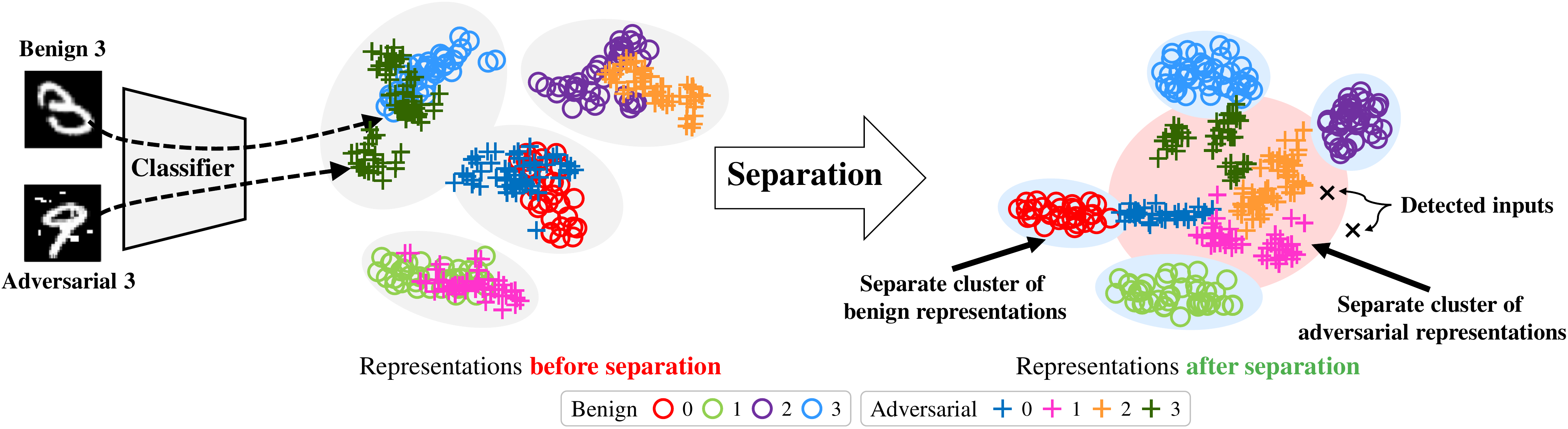}
  \caption{Benign (circles) and adversarial ($+$) representations from a classifier.
  We observe that in the representation space, the adversarial distribution overlap
  with the benign distribution before our defense method is used (Left).
  After our method is used, adversarial examples are well separated from benign examples facilitating the robust detection of attacks (Right).}
  \label{fig:vfr-illustration}

\end{figure*}

We describe our idea in~\autoref{fig:vfr-illustration}
which shows the representations of benign and adversarial examples. 
We prepare a pre-trained deep neural network that classifies MNIST\cite{mnist} 
dataset of hand-written digits images.
We generate adversarial examples on the classifier and visualize
the representations of them reducing its dimension in 2D plane.
In \autoref{fig:vfr-illustration} (left), 
the circles are the representations of benign examples
and the plus sign indicate the representations of adversarial examples. 
Colors denote the predicted labels.
Contrary to the popular belief that adversarial examples are
random samples, representations of
adversarial examples form their own clusters (clusters of $+$ with same colors), 
rather than surrounding benign clusters (clusters of circles with same colors).
Furthermore, for each label, its benign and adversarial clusters overlap each other, which  
makes it difficult to detect adversarial examples from their representations.
It is therefore desirable that adversarial and benign examples from separate clusters in the representation space such that we can use distance-based measures (or equivalently, likelihood-based measures) to detect adversarial examples  (\autoref{fig:vfr-illustration} right, black $\times$). 
To achieve this, we propose a deep encoder which 
maximizes class conditional likelihood assuming 
Gaussian mixture model in the  representation space.
We introduce an additional cluster for the adversarial representations,
and also maximize its class conditional likelihoods. 

To the best of our knowledge, it is the first attempt in the literature
that adversarial attacks are explained by overlapping distributions of representations and consequently addressed by separating the clusters.
Existing detectors focus only on \textit{clusters of benign representations}~\cite{simple,rethinking}
and do not attempt to model adversarial representations.
Similar to our idea of clusters of adversarial representations,
a prior work~\cite{nrf} defines a common property of adversarial examples
as a mathematical function but does not suggest a defense method. 

By learning the \textit{separate} clusters of benign and adversarial representations, we encourage a neural network to learn distinct representations of benign examples that are clearly distinguished from adversarial representations, which enables robust adversarial detection based on distances in the representation space.

We summarize our contributions as follows.
\begin{itemize}
    \item We make new observations that adversarial examples present themselves as clusters in the representation space that overlap with benign clusters, and posit that it poses a fundamental limitation on the detection of adversarial examples in the previous defense methods. 
    \item We propose a new robust adversarial detector 
    and a training algorithm for it which effectively detects
    adversarial examples under an adaptive whitebox threat model by estimating separate
    clusters of adversarial representations.
    \item We conduct comprehensive experiments against 7 adaptive whitebox 
    and blackbox attacks with 4 different perturbation types.
    Our approach improves the worst-case detection performance (Attack Success Ratio) by up to 45\% (MNIST) and 27\% (CIFAR10) compared to the state-of-the-art detectors.
\end{itemize}



\section{Background}

\subsection{Neural network}
Deep Neural Network (DNN) is a function, $H_\theta:\mathbf{X} \rightarrow \mathbf{Z}$,  
which maps $n$ dimensional input space $\mathbf{X}$ to $m$ dimensional output space $\mathbf{Z}$.
In general, DNN is a composition of many sub-functions $h_i$ which are also called layers,
such that $H_\theta(\mathbf{x}) = h_L\circ h_{L-1} \circ \cdots \circ h_1(\mathbf{x})$.
The popular examples of layers include fully connected layers and convolutional layers
followed by activation functions such as ReLU, sigmoid or linear.
\textbf{Representation}s are outputs of each layer, $\cup_{i=1}^L h_i \circ h_{i-1} \cdots \circ h_1(\mathbf{x})$, given an input $\mathbf{x}$.
In this paper, our main interest is the representation of the last layer $H_\theta(\mathbf{x})=h_L \circ h_{L-1} \cdots \circ h_1(\mathbf{x})$.

For k-class classification tasks, practitioners build a classifier $f:\mathbf{X} \rightarrow \mathbf{Y}$
multiplying $k \times m$ matrix  $\mathbf{W}$
to $H_\theta(\mathbf{x})$ so that $g(\mathbf{x})=\mathbf{W}H_\theta(\mathbf{x})=[g_1(\mathbf{x}), g_2(\mathbf{x}), \cdots, g_k(\mathbf{x})]^T$.
Then, $g_i(\mathbf{x})$ are normalized by softmax function
and becomes $f_i(\mathbf{x}) = e^{g_i(\mathbf{x})}/\sum_{j=1}^k e^{g_j(\mathbf{x})}$.
We can interpret $f_i(\mathbf{x})$ as a classification probability of $\mathbf{x}$ to label $i$, $p(y=i|\mathbf{x})$ 
because, $f_i(\mathbf{x})$ ranges from 0 to 1 and $\sum_{i=1}^kf_i(\mathbf{x}) = 1$.
Then, classification label of $\mathbf{x}$ is $\hat{y}= \argmax_{i} f_i(\mathbf{x})$. 

To train the classifier with training data $\mathcal{D}$,
we compute loss function $\mathcal{L}(f(\mathbf{x}), y), (\mathbf{x},y) \in \mathcal{D}$,
which indicates how the classification differs from the correct label $y$.
Then we update parameters $\theta$ and $W$ so that resulting $f$ minimizes the loss.
In particular, we use gradient descent method with back-propagation algorithms.
\subsection{Adversarial evasion attack}
Adversarial evasion attacks modify an input $\mathbf{x}$
by adding imperceptible perturbation $\delta$ in test time,
so that $\mathbf{x}+\delta$ is misclassified to a wrong label, although the original $\mathbf{x}$ is classified correctly.
Formally, the main objective of attacks is to find $\delta$ satisfying,
\begin{gather}
y = \argmax_if_i(\mathbf{x}) \land  y \neq \argmax_if_i(\mathbf{x} +\delta)\\
s.t. \; ||\delta||_p \leq \epsilon, \,\, (\mathbf{x}, y) \in \mathcal{D}
\end{gather}
The first statement implies the condition of misclassification of perturbed image $\mathbf{x} + \delta$, 
and the second statement constraints the size of the perturbation $\delta$ to be smaller than $\epsilon$ so that the original semantics of $\mathbf{x}$ is not changed.  $||\delta||_p$ is called $p$-norm or $\ell_p$ of $\mathbf{\delta}$, and it is computed as,
$$
||\delta||_p=\sqrt[\leftroot{-1}\uproot{10}p]{\sum_i |\delta_i|^p}
$$
Depending on attack constraints and semantics of $\mathbf{x}$, 
the representative $p$ values can be 0, 1, 2 or $\infty$.
In particular $\ell_0$ of $\delta$ restricts the number of non-zero elements,
and $\ell_\infty$ of $\delta$ restricts $\max_i|\delta_i|$,
the maximum value of changes.

Since a first discovery of adversarial attack \cite{szegedy},
many variants of evasion attacks are developed.
They can be categorized
into two classes depending on information adversaries can utilize 
in finding $\delta$, whitebox attacks and blackbox attacks.
\paragraph{Whitebox attack.} In whitebox attack model, adversary can access the
victim classifier $f(\mathbf{x})$ and observe the output and representations of $f(\mathbf{x})$.
In particular, whitebox attacks leverage backpropagation algorithm to compute
gradient of classification loss, $\nabla_{\mathbf{x}} \mathcal{L}(f(\mathbf{x}), y)$ and
modify $\mathbf{x}$ based on the gradient.

$\bullet$ \textbf{Fast Gradient Sign Method (FGSM)} \cite{fgsm} is a basic whitebox attack method which generates 
an adversarial example as,
$$
\mathbf{x_{\text{adv}}} = \mathbf{x} + \epsilon\, \text{sign}(\nabla_{\mathbf{x}} \mathcal{L}(f(\mathbf{x}), y))
$$
FGSM increases the classification loss by adding $\delta = \epsilon\,\text{sign}(\nabla_{\mathbf{x}} \mathcal{L}(f(\mathbf{x}), y))$ to the original $\mathbf{x}$ so that $\mathbf{x_{\text{adv}}}$ is misclassified to wrong label other than $y$. By taking sign of gradient, FGSM satisfies $||\delta||_\infty = ||\epsilon\, \text{sign}(\nabla_{\mathbf{x}} \mathcal{L}(f(\mathbf{x}), y))||_\infty= \epsilon \leq \epsilon$.

$\bullet$ \textbf{Projected Gradient Descent (PGD)} is a stronger attack than FGSM which can be
considered as applying FGSM multiple times.
PGD attack \cite{pgd} has three attack parameters, $\epsilon$, $\alpha$ and $k$.
The parameter $\epsilon$ constraints the size of $\delta$, and
$\alpha$ is a size of single step perturbation of FGSM.
$k$ is the number of application of FGSM attack.
PGD can be represented as follows.
\begin{gather}
    \label{eq-pgd-attack}
    \mathbf{x^{t+1}} =\text{clip}_{(\mathbf{x}, \epsilon)}( \mathbf{x^{t}}+ \alpha\,\text{sign}(\nabla_{\mathbf{x}} \mathcal{L}(f(\mathbf{x^{t}}), y))\\
    \mathbf{x^0} = \mathbf{x}\,\\ \mathbf{x}_\text{adv} = \mathbf{x}^k
\end{gather}
The sign function is an element-wise function which is 
1 if an input is positive otherwise -1.
$\text{clip}_{(\mathbf{x}, \epsilon)}(\mathbf{x}')$ is an element-wise function which enforces $||\delta||_{\infty} = ||\mathbf{x}_{\text{adv}} - \mathbf{x}||_{\infty} \leq \epsilon$. It is defined as,
\[
\text{clip}_{(x_i, \epsilon)}(x'_i) =
\begin{cases}
   x'_i &\text{if } |x'_i - x_i|\leq \epsilon\\
    \epsilon\,\text{sign}(x'_i - x_i) & \text{otherwise}
\end{cases}
\]

The above attacks are explained in the perspective of $\ell_{\infty}$ perturbation, however they can be extended to other perturbations $\ell_p$ by replacing the sign and clip function to different vector projection and normalization functions.
Also attacks can induce a specific target wrong label (targeted attack) rather than arbitrary labels as above (untargeted attack). For target label $y_{\text{target}}$, we replace $\mathcal{L}(f(\mathbf{x^{t}}), y))$ to $\mathcal{L}(f(\mathbf{x^{t}}), y_{\text{target}}))$ and negate the sign of perturbation to minimize the loss,
\begin{gather}
    \mathbf{x^{t+1}} =\text{clip}_{(\mathbf{x}, \epsilon)}( \mathbf{x^{t}} - \alpha\,\text{sign}(\nabla_{\mathbf{x}} \mathcal{L}(f(\mathbf{x^{t}}), y_{\text{target}}))
    \label{eq:targeted-pgd-attack}
\end{gather}

Especially, we call a circumstance \textbf{adaptive whitebox attack model} if an adversary also knows victim's defense mechanism in addition to the classifier $f(\mathbf{x})$.
In the adaptive whitebox attack model, attacks can be adapted to a specific defense by
modifying their objective function $\mathcal{L}(f(\mathbf{x^{t}}), y)$ as suggested by a prior work \cite{easily}. For instance, assume a defense detects adversarial examples if its defense metric $q:\mathbf{X} \rightarrow \mathbb{R}$ is too low compared to benign data $(\mathbf{x}, y)\in \mathcal{D}$.
Then, an attack can be adapted to the defense by adding $q(\mathbf{x})$ to the base objective function $\mathcal{L}(f(\mathbf{x^{t}}), y)$ so that,
\begin{equation}
\mathcal{L}_{\text{adapt}}(f(\mathbf{x^{t}}), y) = \mathcal{L}(f(\mathbf{x^{t}}), y) + q(\mathbf{x})    
\label{eq-adaptive-attack}
\end{equation}
It is important to evaluate defense mechanisms in
an adaptive whitebox attack model, although the adaptive whitebox attack 
model is not realistic assuming adversary's complete access to victims, because it evaluates 
 robustness of the defenses more thoroughly.

\paragraph{Blackbox attack.}
\label{paragraph-blackbox-attacks}
In blackbox attack model, an adversary only observes
partial information about the victim such as training data, classification probabilities $p(\mathbf{y}|\mathbf{x})$ or classification label $\hat{y}=arg\max_if_i(\mathbf{x})$. It tries to generate adversarial examples based on the limited information.
We introduce representative three blackbox attacks here.

$\bullet$ \textbf{Transfer attack} \cite{delving} only assumes an access to training data $\mathcal{D}$ of a victim classifier $f(\mathbf{x})$.
An adversary trains their surrogate classifier $f_s(\mathbf{x})$ with $\mathcal{D}$ 
and generates adversarial examples $\mathbf{x}_\text{adv}$ on $f_s$. Then the adversary applies $\mathbf{x}_\text{adv}$ to the victim classifier $f(\mathbf{x})$.
It leverages a property of adversarial examples called \textit{transferability} which means adversarial examples generated from a classifier successfully induce wrong labels on other classifiers in high probability.

$\bullet$  \textbf{Natural Evolution Strategy (NES)}~\cite{nes} assumes an access to classification probabilities of an input, $f(\mathbf{x})$. Based on the classification probabilities of $n$ neighbor inputs which are sampled by Gaussian distribution $N(\mathbf{x}, \sigma\mathbf{I})$ where $\mathbf{I}$ is an identity matrix, it estimates gradient of expected classification probability as follows,
$$
\nabla_{\mathbf{x}} \mathbb{E}[f_i(\mathbf{x})] = \frac{1}{\sigma n} \sum_{i=1}^n f_i(\mathbf{x}_i),\; \mathbf{x}_i \sim N(\mathbf{x}, \sigma\mathbf{I})
$$
Then, it follows the same attack process with whitebox attack replacing the backpropagated gradient $\nabla_{\mathbf{x}} \mathcal{L}(f(\mathbf{x^{t}}), y))$ with the estimated gradient $\nabla_{\mathbf{x}} \mathbb{E}[f_i(\mathbf{x})]$ to increase classification probability of $\mathbf{x}$ to a label $i$.

$\bullet$  \textbf{Boundary attack}~\cite{boundary} only assumes an access
to a classification label, $\hat{y} = arg\max_if(\mathbf{x})$.
In contrast to other attacks find $\delta$
starting from $(\mathbf{x}, y) \in \mathcal{D}$, boundary attack starts from a uniform sample in valid input space where $y\neq arg\max_if(\mathbf{x}_0)$.
In a nutshell, after finding $\mathbf{x}_0$, the attack iteratively finds $\mathbf{x}_{t+1}$ in the neighborhood of $\mathbf{x}_t$, 
so that $y\neq arg\max_if(\mathbf{x}_{t+1})$ is still satisfied while minimizing $||\mathbf{x} - \mathbf{x}_{t+1}||_2 < ||\mathbf{x} - \mathbf{x}_{t}||_2$. After $k$ iterations, $\mathbf{x}_k$ approaches very close to $\mathbf{x}$ while classified to wrong label.


\subsection{Threat model}
\label{threatmodel}
\paragraph{Problem and threat model.} A problem we deal with in this paper is to find
a neural network model which robustly detects adversarial examples so that main classification tasks
are not compromised by adversaries.
As a metric of the detection performance, we define \textbf{Attack Success Ratio} $\mathcal{S}(f,q)$, 
similar to a prior work~\cite{trade} (lower is better):
\begin{gather}
\begin{aligned}
\mathcal{S}(f,q) = \mathbb{E}_{(\mathbf{x}, y)\sim \mathcal{D}}&[\textbf{1}\{\exists{||\delta||_p \leq \epsilon} \\
&\mbox{s.t.}\;f(\mathbf{x}+\delta)\neq f(\mathbf{x}) = y,\;q(\mathbf{x}) \leq 0\}]
\end{aligned}
\end{gather}
$f$ is a classifier which outputs a class label and $q$ is a detector which is greater than 0
if an input is adversarial. 
$\mathbf{1}\{\textit{cond}\}$ is 1 if \textit{cond} is true, otherwise 0.
So $\mathcal{S}(f,q)$ is a proportion of $(\mathbf{x}, y) \in \mathcal{D}$ 
where adversaries can find $\mathbf{\delta}$ which misleads $f$ and bypasses $q$
, over the entire data points in $\mathcal{D}$.
Therefore our objective is to find neural networks lowering $\mathcal{S}(f,q)$.

We assume \textbf{adaptive whitebox adversaries} who know everything of victim models and defense mechanisms.
It is an attack model suggested in prior works \cite{biggio2013evasion, easily}
where we can analyze effective robustness of a defense method.
For example, adversaries can compute gradients to generate adversarial examples, and see intermediate values of victim models.
The objective of the adversaries is to fool our detector and classification, increasing $\mathcal{S}(f,q)$.
On the other hand, defenders can change the parameters of victim models to resist against the adversaries \textit{in a training phase}. 
However, \textit{after the training phase}, the defenders do not have control of the victim models so it is the most challenging condition to defenders.
We also consider blackbox attacks where the adversaries do not have knowledge of the victim models to ensure that our defense does not depend on obfuscated gradients~\cite{obfuscated}.
\section{Approach}
In this section, we propose an adversarial detector
against adaptive whitebox adversaries. 
First, we empirically demonstrate \textbf{clusters of adversarial 
representations} in neural networks (Section 3.1).
In addition, we find the adversarial clusters
highly overlap with benign representations,
which makes it hardly possible for any existing detectors 
to detect them based on the representations.
Motivated by the fundamental limitation,
we introduce a new adversarial detector and a training method (Section 3.2) 
which \textbf{separates the adversarial representations in a
new cluster} to minimize the overlaps with benign representations.
To the best of our knowledge, it is the first time to identify adversarial 
examples as clusters of representations and deal with them
by cluster separations.
For detailed comparisons to existing approaches, we refer readers to 
the related work section (Section 4).

\subsection{Motivation}

\label{motivation_section}
In this section, we demonstrate an experiment 
where we identify clusters of adversarial representations
and get motivated to separate them from benign representations.
We prepare MNIST\cite{mnist} training data $\mathcal{D}$ which contains hand-written
digit images form 0 to 9 ($k=10$ classes), and a deep neural network
$H_\mathbf{\theta}:\mathbf{X} \rightarrow \mathbb{R}^{k}$ initialized with parameters $\mathbf{\theta}$.
$H_\mathbf{\theta}$ maps $\mathbf{x}$ to a \textit{representation} $\mathbf{z}$ in $k$ dimensional space.
We also assign $\mathbf{\mu}_0, \mathbf{\mu}_1, \cdots, \text{and}\; \mathbf{\mu}_{k-1}$
in the representation space, and let $\mathbf{\mu}_i$ be a center of
representations $H_\theta(\mathbf{x}_{y \Equal i})$, 
where $\mathbf{x}_{y=i}$ is defined as $\mathbf{x}$ whose label is $i$, 
$(\mathbf{x}, y=i) \in \mathcal{D}$.
In particular, we let $\mu_i$ be an one hot encoding 
of label $i$ as similar with a prior work~\cite{resisting}.

We train $H_{\theta}$ by updating the parameters $\mathbf{\theta}$ so that $H_\theta(\mathbf{x}_{y=i})$ is
clustered around $\mathbf{\mu}_i$ with a training objective,
\begin{gather}
\hat{\theta}=\argmin_{\theta}\mathcal{L}(\theta; \mathcal{D}) \\
\mathcal{L}(\theta; \mathcal{D}) =\sum_{(\mathbf{x}, y) \in \mathcal{D}} ||H_{\mathbf{\theta}}(\mathbf{x}) - \mathbf{\mu}_y ||_2
\end{gather}
Optimizing above loss, we can find new parameters $\hat{\theta}$
that minimize Euclidean distance between $H_{\hat{\theta}}(\mathbf{x}_{y\Equal i})$ and $\mathbf{\mu}_i$.
Then, we can devise a classifier $f(\mathbf{x})$ based on the distance,
$$
\hat{y} = f(\mathbf{x})=\argmin_{c} ||H_{\hat{\mathbf{\theta}}}(\mathbf{x}) - \mathbf{\mu}_{c}||_2
$$
\begin{figure}
\centering
\includegraphics[width=7.5cm]{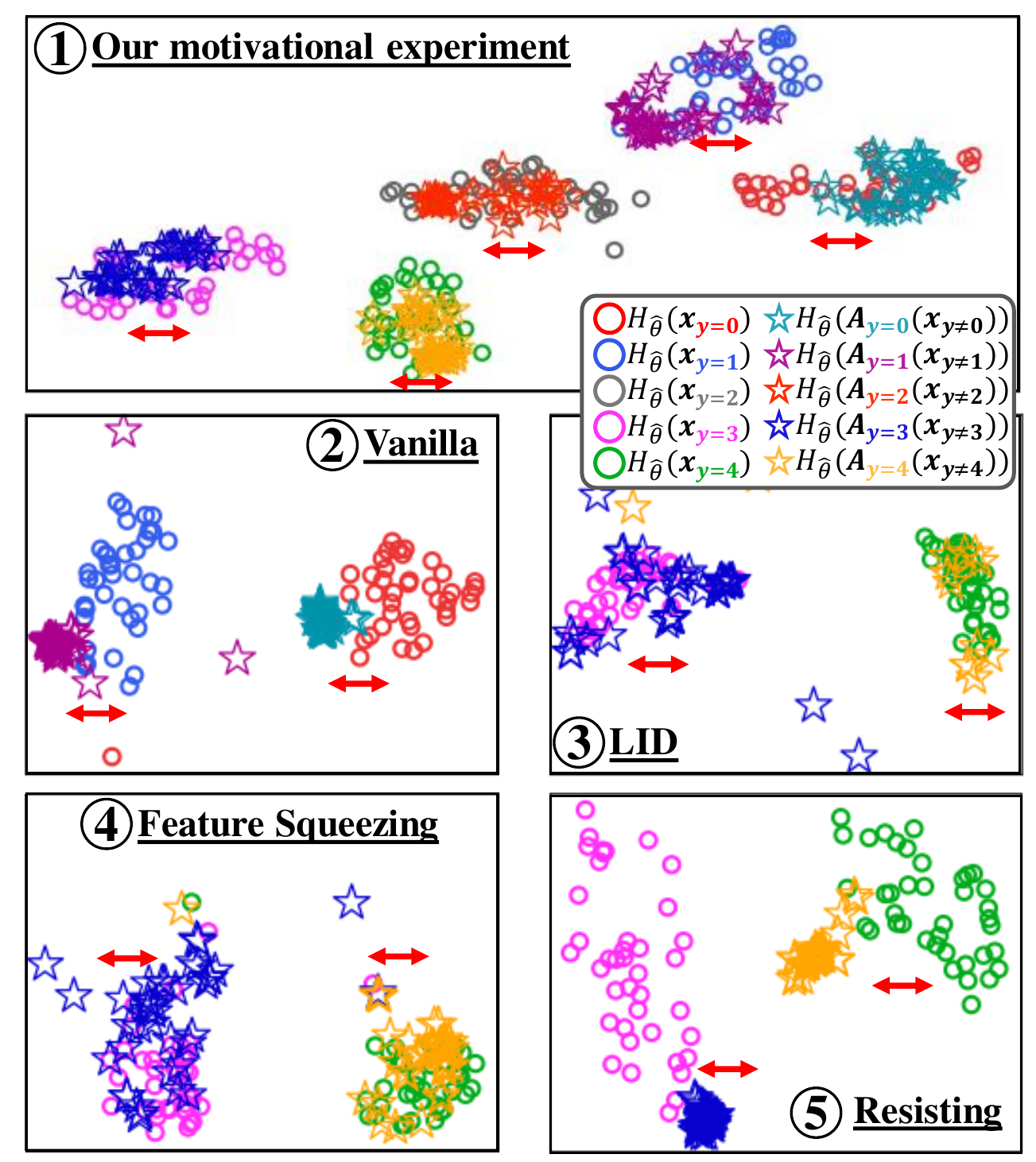}
    \caption{Overlaps of benign and adversarial clusters in a representation space of our motivational experiment (1). 
    We also find the overlaps in representation spaces of a vanilla classifier (2) and other existing detectors (3 $\sim$ 5).  
    }
  \label{fig:overlaps-in-existing-defenses}
\end{figure}
After $H_{\hat{\theta}}$ is trained, we inspect representations of benign and adversarial examples.
To visualize benign representations $H_{\hat{\theta}}(\mathbf{x}), (\mathbf{x}, i) \in \mathcal{D}$, 
we reduce the dimension of $H_{\hat{\theta}}(\mathbf{x})$ to 2 dimensional space 
with t-SNE~\cite{tsne} algorithm.
We plot the result in \circled{1} of \autoref{fig:overlaps-in-existing-defenses}.
We can check that $H_{\hat{\theta}}(\mathbf{x}_{y=i})$ of the same class 
(circles with the same colors)
 are in the same cluster as we intended.
Then we observe representations of adversarial examples $H_{\hat{\theta}}(\mathbf{A}_{{y=i}}(\mathbf{x}_{y\neq i}))$.
$\mathbf{A}_{y=i}$ is an targeted attack algorithm which finds $\mathbf{\delta}$ satisfying $i=f(\mathbf{x}+\delta)$.
We can derive $\mathbf{A}_{y=i}$ by modifying PGD 
attack (\autoref{eq:targeted-pgd-attack}) replacing the attack loss $\mathcal{L}$ 
with $||H_{\mathbf{\hat{\theta}}}(\mathbf{x} + \mathbf{\delta}) - \mathbf{\mu}_i||_2$.
The adversarial examples are generated with $\ell_{\infty}$ 0.3 perturbations and
$H_{\hat{\theta}}(\mathbf{A}_{{y=i}}(\mathbf{x}_{y\neq i}))$ are also plotted as stars in \circled{1} of \autoref{fig:overlaps-in-existing-defenses}.
Not only the adversarial representations  $H_{\hat{\theta}}(\mathbf{A}_{y\Equal i}(\mathbf{x}_{y\neq i}))$
are close to centers $\mathbf{\mu}_i$ of the misleading target class $i$, but \textbf{the adversarial representations
also forms their own clusters near $\mathbf{\mu}_i$}.
The existence of the cluster of adversarial representations is not expected,
because the adversarial examples $\mathbf{A}_{{y=i}}(\mathbf{x}_{y\neq i})$ 
are generated from $\mathbf{x}_{y\neq i}$ of various classes $y\in\{0,1,\cdots,k-1\} - \{i\}$ which do
not share visual similarity of class $i$.
We also conducted similar experiments on \circled{2} a vanilla classifier 
and different detector approaches (\circled{3} $\sim$ \circled{5}) including LID\cite{lid}, Feature Squeezing\cite{feature-squeeze} Resisting \cite{resisting}.
We find the overlaps consistently exist in the classifier and detection models.
In the perspective of adversarial detection 
we regard these overlaps are fundamental limitations of existing detectors,
because it is impossible to detect $\mathbf{A}_{y\Equal i}(\mathbf{x}_{y\neq i})$ based on the distance 
metric $||\mathbf{A}_{y\Equal i}(\mathbf{x}_{y\neq i}) - \mathbf{\mu}_i||_2$
as long as the overlaps between benign clusters and 
adversarial clusters exist (red arrows) in the representation space.

\begin{figure}
\centering
\includegraphics[width=5.6cm]{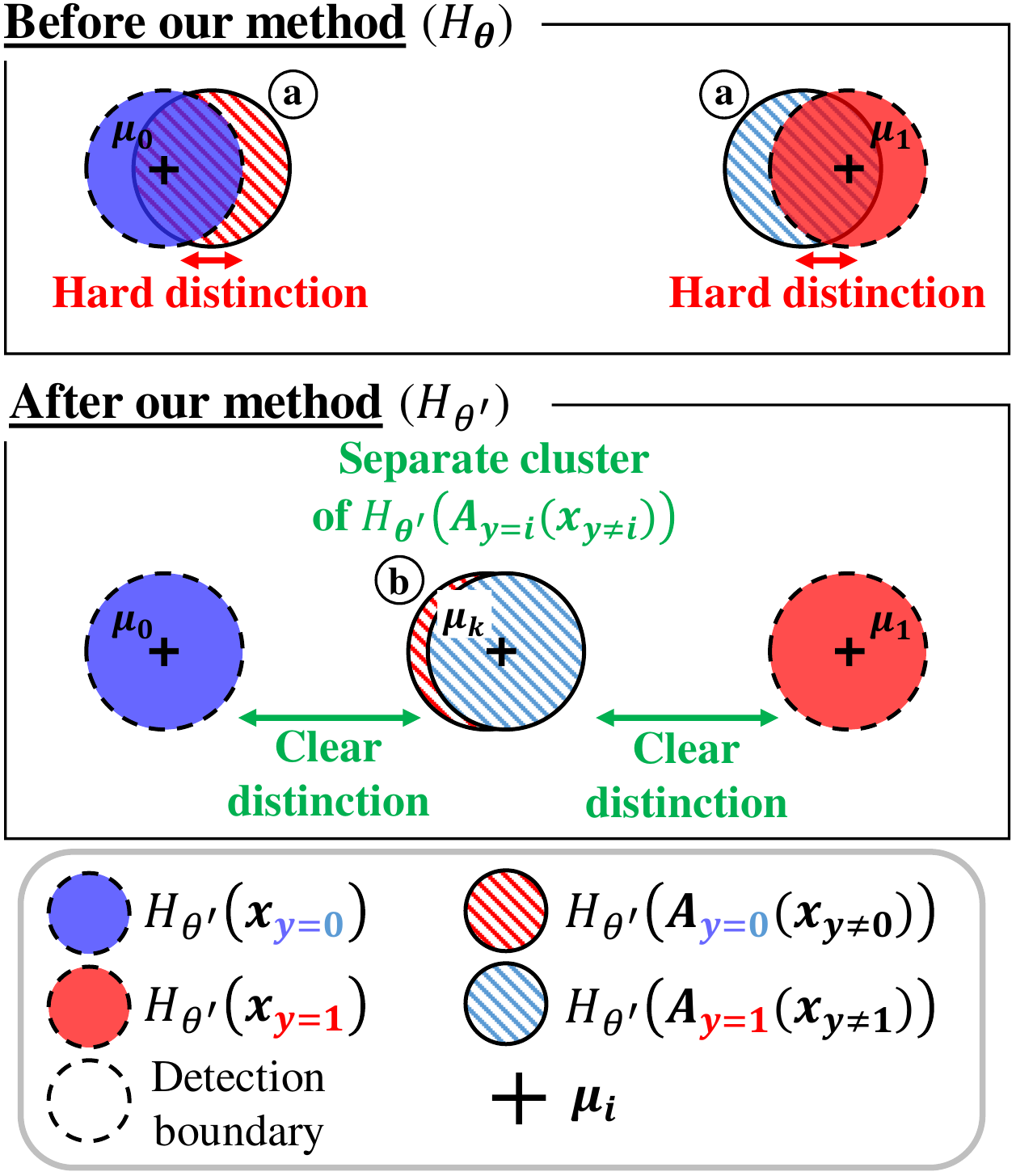}
  \caption{We find overlaps between clusters of adversarial representations
  and clusters of benign representations is a fundamental limitation of existing adversarial detectors
  (Before our method). 
  We solve this separating benign clusters from adversarial clusters for robust adversarial detection.}
  \label{fig:motivation}

\end{figure}

Therefore, we are motivated to find $H_\theta$ which separates benign clusters 
clearly distinguished from the adversarial representations
(red and blue circles surrounded by dashed lines in ``After our method'' of \autoref{fig:motivation}),
so that distances between $\mathbf{\mu}_i$ and $H_{\theta}(\mathbf{x})$ can be effective metrics for 
detecting adversarial examples.
To ensure the non-overlapping clusters, 
we focus on \textbf{the existence of clusters of adversarial representations}  (\circled{a} of \autoref{fig:motivation}).
If the adversarial representations truly form their own clusters, we deduce
it is also possible to find $H_\theta$ that clusters the adversarial examples
in a different location on the representation space.
So we are motivated to generate adversarial examples $\mathbf{A}_{y=i}(\mathbf{x}_{y\neq i})$
in a training process and estimate $\theta'$ which clusters
adversarial representations to a separate center $\mathbf{\mu}_{k}$ (\circled{b} of \autoref{fig:motivation}).

It is our main contribution to identify 
 \circled{a}, \textbf{adversarial examples as a cluster} in the representation space,
and \circled{b}, \textbf{separate the adversarial
representations in a cluster} at $\mathbf{\mu}_{k}$. 
Considering a cluster in the representation space
as a property of $\mathbf{x}$ in the input space,
 clustering the adversarial representations at a center $\mu_k$ 
implies $H_{\theta'}$ captures common properties of adversarial examples
in the input space.
Further separating clusters of adversarial representations
from the cluster of benign representations (\circled{b} of \autoref{fig:motivation}),
we claim the resulting clusters of benign representations reflect
 much distinct and robust properties of the benign examples compared 
 to clusters before the separation. 
Note that the clustering representations of adversarial examples
 at $\mu_k$ is necessary to make sure the separation,
because we can not measure how clearly two clusters are
separated knowing only one of them.
We support above claim in experiments where 
our detectors outperform
existing approaches which do not consider adversarial clusters~.

\subsection{Model construction} 
\begin{figure}
\centering
\includegraphics[width=6.0cm]{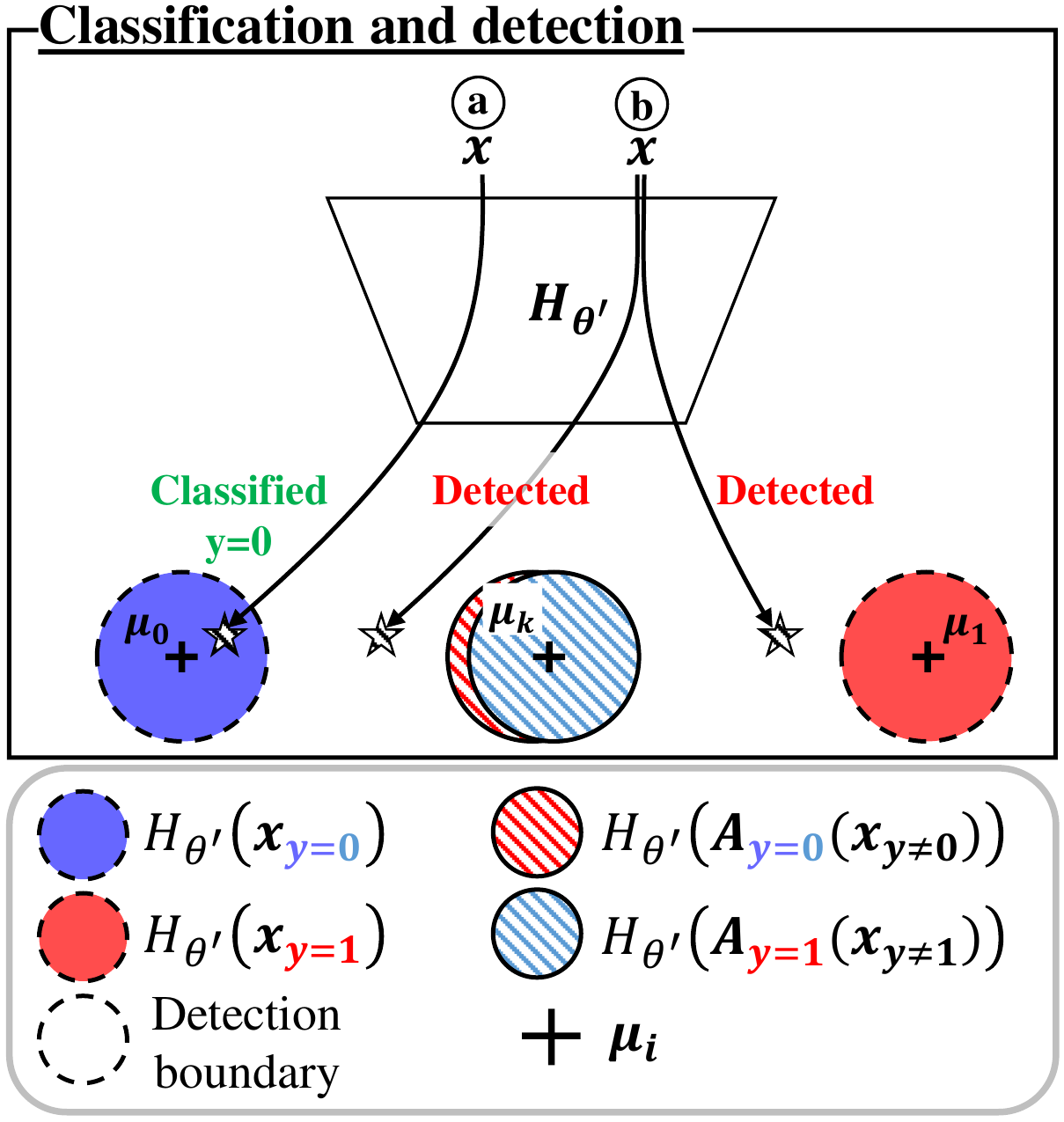}
\caption{Classification and detection method of $H_\theta$.
It classifies an input to a class of the closest Gaussian mode (a),
or detect if the representation of input is located far from 
all of the benign modes (b).}
\label{fig-method-classification-detection}
\end{figure}

For a clear distinction between benign and adversarial representations, we introduce a probabilistic detection model and a training algorithm, which enables likelihood based robust adversarial detection.
We leverage an encoder model $H_\theta:\mathbb{R}^n \rightarrow \mathbb{R}^m$ which is a deep neural network such as ResNet~\cite{resnet} and DenseNet~\cite{densenet}
except that $H_\mathbf{\theta}$ is not forced to produce outputs with $m=k$ dimensions.
To interpret the distance $||H_{\theta}(x) - \mathbf{\mu}_i||_2$ from the motivational experiment
in a probabilistic view,
we introduce $k$ Gaussian mixture model (GMM) in the representation
space which contains $k$ Gaussian modes whose means and variances are parameterized with class label $y \in \{0, 1, \cdots, k-1\}$:
\begin{gather}
\begin{aligned}
p(H_\theta(\mathbf{x})) &= \sum_{i=0}^{k-1}p(y\Equal i)p(\mathbf{z}\Equal H_\theta(\mathbf{x})|y\Equal i)\\
& = \sum_{i=0}^{k-1}  p(y\Equal i) N(H_\theta(\mathbf{x});\mathbf{u}(i), \mathbf{v}(i))
\label{eq-gmm}
\end{aligned}
\end{gather}
$p(y\Equal i)$ is a probability distribution of class labels $y$ of $\mathcal{D}$.
In our case we can assume $p(y\Equal i)$ to be equiprobable, $\forall i \in \{0,1,\cdots, k-1\}, p(y\Equal i)=\frac{1}{k}$, because our datasets include the same number of $\mathbf{x}$ over classes.
The modes of Gaussian mixture model are
Gaussian distributions $N(H_\theta(\mathbf{x})|\mathbf{u}(i), \mathbf{v}(i))$
whose mean $\mu_i$ is $\mathbf{u}(i)$ and variance $\sigma^2$ is $\mathbf{v}(i)$.
The modes correspond to the clusters in the motivational experiment
and they are illustrated with blue and red circles surrounded by dashed lines in \autoref{fig-method-classification-detection}.
We properly choose parameterizing function $\mathbf{u}$ and $\mathbf{v}$ so that Bayes error rates between the Gaussian modes should be low. Specifically, 
\begin{gather}
l = \frac{m}{k} \in \mathbb{N} \\
u_j(y) = 
\begin{cases}
   M &\text{if  } yl \leq j < (y+1)l\\
    0 & \text{otherwise}
\end{cases}\\
\mathbf{v}(y) = \mathbf{I}_m
\end{gather}
$\mathbf{u}:\mathbb{R} \rightarrow \mathbb{R}^m$ returns $m$ dimensional vector given a class
label $y$, whose $j^{\text{th}}$ element is defined by $u_j(y)$.
In other words, $\mathbf{u}(y)$ returns a vector whose $l$ consecutive elements are constant $M$
starting from $yl^{\text{th}}$ element, otherwise 0.
$\mathbf{u}$ can be considered as an extension of one hot encoding (if $m=k, M=1$)
to be applied 
to $m=kl$ dimensional representations. 
We let $\mathbf{v}$ be a constant function which returns $m\times m$ identity matrix $\mathbf{I}_m$, for
the efficient computation from isotropic Gaussian distributions.

From the GMM assumption, we can derive the distance $||H_{\theta}(\mathbf{x})-\mu_i||_2$ re-writing $N(H_\theta(\mathbf{x});\mathbf{u}(i), \mathbf{v}(i))$
in \autoref{eq-gmm}:
\begin{gather}
\begin{aligned}
p(H_\theta(\mathbf{x})) 
& = \sum_{i=0}^{k-1}  p(y\Equal i) N(H_\theta(\mathbf{x});\mathbf{u}(i), \mathbf{v}(i)) \\
& = \sum_{i=0}^{k-1}  \frac{1}{k} \frac{1}{(2\pi)^{n/2}}e^{-\frac{1}{2}||H_\theta(x_{i})-\mu_{y_{i}}||^2_2}
\end{aligned}
\end{gather}
Similar construction is introduced in prior works \cite{simple,resisting,rethinking} and shown effective in classification tasks with flexible transformation of $H_\theta$.
Now we can compute likelihood of the representations $p(H_{\theta}(\mathbf{x}))$ with the distances $||H_\theta(x_{i})-\mu_{y_{i}}||_2$.

\paragraph{Classification and Detection.} \label{paragraph-detection} Assuming $H_{\theta'}$ 
successfully 
clusters benign and adversarial representations in separate Gaussian modes
(we explain a training method of $H_{\theta'}$ in the following section)
we can classify inputs according to Bayes classification rule, and
the GMM assumption of $H_{\theta'}$:
\begin{gather}
\begin{aligned}
f(\mathbf{x}) &= \argmax_y p(y|z\Equal H_{\theta'}(\mathbf{x})) \\
            & = \argmax_y \frac{p(z\Equal H_{\theta'}(\mathbf{x})|y)p(y)}{p(z\Equal H_{\theta'}(\mathbf{x}))}\\
     &= \argmax_y p(z\Equal H_{\theta'}(\mathbf{x})|y) \\
     & = \argmax_{y} N(z\Equal H_{\theta'}(\mathbf{x});\mu_{y}, \mathbf{I}_m) \\     
    & = \argmin_{y}||H_{\theta'}(\mathbf{x})-\mu_{y}||_2
\end{aligned}
\end{gather}

After all, $\mathbf{x}$ is classified to a class label $y$ where a distance 
$||H_{\theta'}(\mathbf{x}) - \mathbf{\mu}_y||_2$ is minimized
(\autoref{fig-method-classification-detection} \circled{a}).
The first equation is from Bayes classification rule and 
re-written with Bayes rule in the second equation.
The $p(y)$ and denominator $H_\theta(\mathbf{x})$ are disappeared in the third equation 
because they are invariant to $y$.
The conditional likelihood $p(z\Equal H_\theta(\mathbf{x})|y)$
is re-written as a Gaussian mode $N(z\Equal H_\theta(\mathbf{x});\mu_{y}, \mathbf{I}_{m})$
in the fourth equation.
Following equation is reduced by ignoring constant of Gaussian distribution
and the $\argmax_y$ is changed to $\argmin_y$ negating the sign of the exponent of $e$
in Gaussian distribution.

We explain how $H_{\theta'}$ detects an input $\mathbf{x}$ as an adversarial examples.
It is illustrated in \autoref{fig-method-classification-detection} \circled{b}.
We measure a distance $||H_{\theta'}(\mathbf{x}) - \mathbf{\mu}_{\hat{y}}||_2$, where
$\hat{y}$ is a classified label,
$\hat{y} = \argmin_{y}||H_{\theta'}(\mathbf{x})-\mu_{y_{i}}||_2$.
Then we compare the $||H_{\theta'}(x) - \mathbf{\mu}_{\hat{y}}||_2$ with 
a threshold $\tau_{\hat{y}}$ (denoted as a radius of a circle surrounded with dashed line in \autoref{fig-method-classification-detection}). 
If $||H_{\theta'}(x) - \mathbf{\mu}_{\hat{y}}||_2$ is greater than $\tau_{\hat{y}}$
we detect $\mathbf{x}$ as an adversarial examples, otherwise
we admit to classify $\mathbf{x}$ as $\hat{y}$.
We have $k$ thresholds of $\tau_i$ for each class and we set $\tau_i$ so that
there are 1\% false positive benign example of class $i$.
In other words, there are benign examples which are
falsely detected as adversarial examples, and their proportion over
entire benign examples is 1\%.



\subsection{Training algorithm}
 
    

\begin{figure}
\centering
\includegraphics[width=6.5cm]{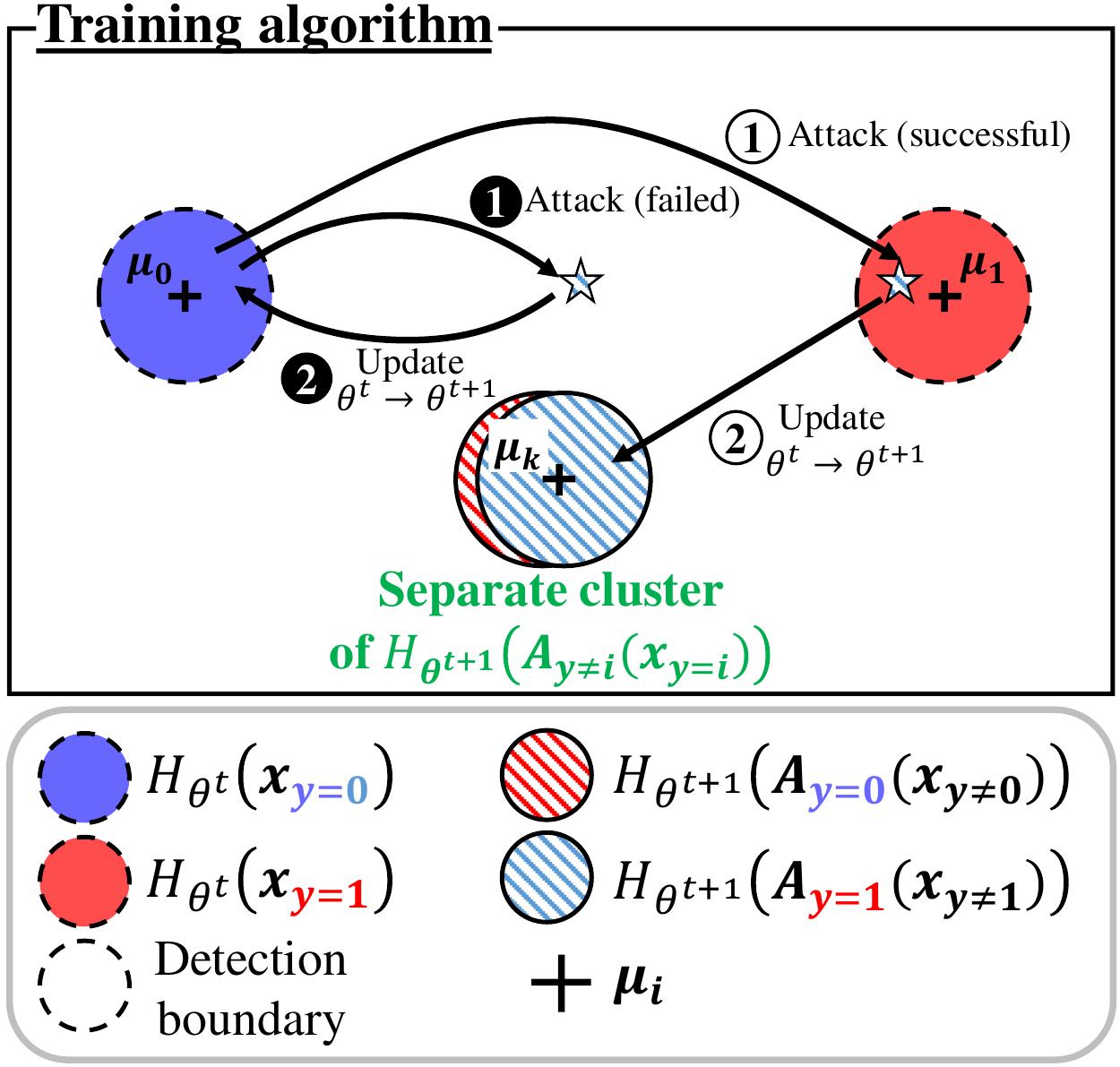}
\caption{Training algorithm of our detector. 
The successful adversarial examples are 
re-labeled as $y=k$, and the failed adversarial examples
are re-labeled as its original class before the attack.
We update $\theta_t$ so that distances between representations 
and centers of their cluster are minimized.}
\label{fig-method-training-test}
\end{figure}

To achieve a robust adversarial detection given the detection and classification model
, we propose the following training algorithm as illustrated in \autoref{fig-method-training-test}.
The training algorithm is a repetitive process 
where its iteration is composed of two phases.
The first phase is for generating adversarial examples and
the second phase is for updating model parameter $\theta$.
\paragraph{Phase 1. Generation of adversarial examples.}
This phase corresponds to \circled{1}, \bcircled{1} of \autoref{fig-method-training-test}.
Before we cluster adversarial representations,
we need to generate adversarial examples against $H_{\theta^t}(\mathbf{x})$.
with an adaptive attack against $H_\mathbf{\theta^t}$.
For adaptive adversaries to bypass and mislead our model $H_\theta$
it is required to to minimize $||H_{\theta^t}(\mathbf{x})-\mu_i||_2$
because our model depends on the single metric, $||H_{\theta^t}(\mathbf{x})-\mu_i||_2$, 
for both of the detection and classification.
Specifically, the adversaries should satisfy
$||H_{\theta^t}(\mathbf{x})-\mu_i||_2 \leq \tau_i$. 
We adapt an attack modifying the existing PGD attack algorithm
by replacing attack loss $\mathcal{L}$ with $||H_{\theta^t}(\mathbf{x})-\mu_i||_2$.
So the resulting adaptive $\ell_\infty$ PGD attack is:
\begin{gather}
    \mathbf{x^{j+1}} =\text{clip}_{(\mathbf{x}, \epsilon)}( \mathbf{x^{j}} - \alpha\,\text{sign}(\nabla_{\mathbf{x}} ||H_{\theta^t}(\mathbf{x}) - \mathbf{\mu}_i||_2))\\
    \mathbf{x^0} = \mathbf{x}\,\\
    A_{y=i}(\mathbf{x}) = \mathbf{x}^k
\end{gather}
We update $i = \argmin_{i\neq y}||H_{\theta'}(\mathbf{x}^j)-\mu_i||_2$ in every attack step $j$.
We generated adversarial examples with attack parameters ($\epsilon=0.3$, $\alpha=0.01$, $k=40$)
for training $H_\theta$ on MNIST,
and ($\epsilon=0.03$, $\alpha=0.007$, $k=20$) for the case of CIFAR10.
This setting is introduced in a prior work~\cite{pgd} and 
considered as a standard in adversarial training.

Although we adaptively conduct attacks against $H_{\theta^t}$, the
generated adversarial examples are not always successful.
The successful attack is a case where
the adaptive attack finds a perturbation $\mathbf{\delta}$
for $(\mathbf{x}, y\Equal i) \in \mathcal{D}$ 
so that  $H_\theta(\bold{x} + \delta)$ gets inside of
a detection boundary of other class $j\neq i$.
In  \circled{1} of \autoref{fig-method-training-test}, it is illustrated that a successful adversarial
example, whose original label is 0, crosses a detection boundary (dashed lines of circles)
of class $y=1$ with perturbation $\delta$.
On the other hand, a failed attack is a case
where the adaptive attack can not find $\mathbf{\delta}$ such that 
$H_\theta(\bold{x} +\delta)$ does not get inside of 
decision boundaries of all other classes except 
the original label of $\bold{x}$.
It is also illustrated as \bcircled{1} of \autoref{fig-method-training-test}.
The representation of failed adversarial example is not included in detection boundaries of any clusters $\mathbf{\mu}_{j\neq 0}$.
We distinguish these two types of adversarial examples and deal with them in 
different ways in the following phase.

\paragraph{Phase 2. Re-labeling data and updating $\theta$.}
After generating adversarial examples in the phase 1.
We update parameter $\mathbf{\theta}^{t}$ to $\mathbf{\theta}^{t+1}$ with the 
adversarial examples and benign training data $\mathcal{D}$.
We deal with those data with different re-labeling rules 
depending on representation clusters which the data should belong to.
After the re-labeling, we get a new dataset 
$\mathcal{D}^t = \mathcal{D}_S^t \cup \mathcal{D}_F^t \cup \mathcal{D}$.

\textbf{A. Re-labeling successful adversarial examples.} We briefly explained successful adversarial
examples in the description of the phase 1.
They are adversarial examples which mislead our classification and bypass 
 detection mechanism. It is strictly defined as $A_S(\mathcal{D})$,
\begin{equation}
\begin{aligned}
A_S(\mathcal{D})= \{&A_{y=j}(\mathbf{x})|\; (\mathbf{x}, i) \in \mathcal{D}, \\
       & \exists j \neq i\; ||H_{\theta^t}(A_{y=j}(\mathbf{x})) - \mathbf{\mu}_j||_2 \leq \tau_j 
\}
\end{aligned}
\end{equation}
To cluster the representations of  $A_S(\mathcal{D})$
in a separate adversarial cluster (\circled{2} of \autoref{fig-method-training-test}),
we define the dataset $\mathcal{D}^t_S$ by re-labeling
$A_S(\mathcal{D})$ as $y=k$.
\begin{equation}
\mathcal{D}^t_S = \{(\mathbf{x}, k)|\; \mathbf{x} \in A_S(\mathcal{D})\}
\end{equation}
We set $\mathbf{\mu}_k$, a center of adversarial 
representations, to be an origin $\mathbf{0}_m$ in the representation space.

\textbf{B. Re-labeling failed adversarial examples.} Because an attack algorithm $A_{y=j}$
does not guarantee generation of $A_S(\mathcal{D})$, there exist
adversarial examples which are perturbed with some $\mathbf{\delta}$ but
failed to cross detection boundaries of other classes (\bcircled{1} of \autoref{fig-method-training-test}).
We strictly define the failed adversarial examples as $A_F(\mathcal{D})$,
\begin{equation}
\begin{aligned}
A_F(\mathcal{D})= \{&A_{y=j}(\mathbf{x})|\; (\mathbf{x}, i) \in \mathcal{D}, \\
       & \forall j \neq i\; ||H_{\theta^t}(A_{y=j}(\mathbf{x})) - \mathbf{\mu}_j||_2 > \tau_j 
\}
\end{aligned}
\end{equation}
Since representations of the failed adversarial examples $A_F(\mathcal{D})$ neither are close
to a cluster nor they form a cluster, re-labeling $A_F(\mathcal{D})$ to $k$ 
as like $A_S(\mathcal{D})$ has negative effects
in clustering adversarial representations.
Instead we re-label the failed adversarial example $A_F(\mathcal{D})$ 
to their original labels before the attack:
\begin{gather}
\begin{aligned}
\mathcal{D}^t_F = \{(A_{y=j}&(\mathbf{x}), i)|\; (\mathbf{x}, i) \in \mathcal{D}
       ,\\
       &\exists j \neq i \;  A_{y=j}(\mathbf{x}) \in A_F(\mathcal{D})
\}
\end{aligned}
\end{gather}
It is depicted in \bcircled{2} of \autoref{fig-method-training-test}.
By restricting changes of representations from small $\delta$ in input space,
this re-labeling rule makes resulting $H_{\theta^{t+1}}$ more robust to input perturbations.

\textbf{C. Updating model parameter $\mathbf{\theta}$.}
After the re-labeling process, we update parameter $\mathbf{\theta}^t$
to $\theta^{t+1}$ , as illustrated in \circled{2} and \bcircled{2} of \autoref{fig-method-training-test},
to find distinct clusters of benign and adversarial representations.
The updating process can be formalized as a likelihood maximization 
problem on the representations of $\mathcal{D}^t = \mathcal{D}_S^t \cup \mathcal{D}_F^t \cup \mathcal{D}$.
The corresponding training objective can be expressed as follows,
\begin{gather}
    \text{maximize}_{\theta} \mathcal{L}(\mathcal{D}^t;\theta^t) \\
    \mathcal{L}(\mathcal{D}^t;\theta^t) = \sum_{(\mathbf{x}, y) \in \mathcal{D}^t}p(H_{\theta^t}(\mathbf{x})) \label{training-loss-start}
\end{gather}
We further expand \autoref{training-loss-start} under the GMM assumption in the model construction.
\begin{gather}
\begin{align}
    \mathcal{L}(\mathcal{D}^t;\theta^t) &= \sum_{(\mathbf{x}, y) \in \mathcal{D}^t}p(H_{\theta^t}(\mathbf{x}))\\
                                        &= \sum_{(\mathbf{x}, y) \in \mathcal{D}^t} \sum_{i=0}^k p(y\Equal i)p(H_{\theta^t}(\mathbf{x})|y\Equal i) \\
                                        &= \sum_{(\mathbf{x}, y) \in \mathcal{D}^t} p(H_{\theta^t}(\mathbf{x})|y)\\
                                        &=\sum_{(\mathbf{x}, y)\in \mathcal{D}^t} N(z\Equal H_{\theta^t}(\mathbf{x});\mu_{y}, \mathbf{I}_m) \\
                                        & = \sum_{(\mathbf{x}, y)\in \mathcal{D}^t} \mathcal{C} e^{-\frac{1}{2}||H_{\theta'}(\mathbf{x})-\mu_{y}||^2_2}
\end{align}
\end{gather}
If we ignore the positive constant $\mathcal{C}$,
we can regard this training objective minimizes distances 
between $H_{\theta^t}(\mathbf{x})$ and its target clustering center $\mu_y$.
As we update $\theta^{t}$,
$H_{\theta^{t+1}}$ better distinguishes the clusters of adversarial and benign representations.

However, in practice, we should resolve a problem where a ratio between the size of $\mathcal{D}^t_S$
and the size of $\mathcal{D}^t_F \cup \mathcal{D}$ is unbalanced. 
For example, if the size of $\mathcal{D}^t_S$ is 1 and the size of $\mathcal{D}^t_F \cup \mathcal{D}$ is 100,
the effect of $\mathcal{D}^t_S$ to the total loss $\mathcal{L}(\mathcal{D}^t;\theta^t)$ is negligible,
and $H_{\theta^{t+1}}$ hardly clusters the representations of $\mathcal{D}^t_S$.
To fix this problem, we split $\mathcal{D}^t$ into $\mathcal{D}^t_S$ and $\mathcal{D}^t_F \cup \mathcal{D}$
, and compute the sum of average likelihoods of each dataset.
\begin{gather}
\begin{align}
 \mathcal{L}_{\text{fix}}(\mathcal{D}^t;\theta^t) = \frac{\mathcal{L}(\mathcal{D}_S^t;\theta^t)}{|\mathcal{D}_S^t|} + \frac{\mathcal{L}(\mathcal{D}_F^t \cup \mathcal{D};\theta^t)}{|\mathcal{D}_F^t \cup \mathcal{D}|}
\end{align}
\end{gather}
Then we maximize $\mathcal{L}_{\text{fix}}(\mathcal{D}^t;\theta^t)$ instead of $\mathcal{L}(\mathcal{D}^t;\theta^t)$.

After we update the parameter to $\theta^{t+1}$, we go to the phase 1 and repeat this process
until it converges enough, so the size of $\mathcal{D}^t_S$ is not decreasing more.
Specifically we trained our models with the fixed number of epochs.
We used 50 epochs for MNIST  and 70 epochs for CIFAR10 with Adam 
optimizer~\cite{adam} with a learning rate $10^{-3}$.
We provide a pseudo code of stochastic mini-batch training in \autoref{training-algorithm}.
\begin{algorithm}[h]
\begin{algorithmic}[1]
    \STATE{$ t \gets 0 $} 
    \WHILE{$t < \rm{MAX\_EPOCHS} $}
        \STATE{$ B \subset \mathcal{D} $ \tcp{$t^{\text{th}}$ benign mini batch.}} 
        \vspace{0.25em}
        \STATE{\textcolor{white}{.}\tcp{Phase 1. \circled{1}, \bcircled{1} of \autoref{fig-method-training-test}}}
        \STATE{$A_S(B)\gets \{A_{y=j}(\mathbf{x})|\; (\mathbf{x}, i) \in B,\linebreak
        \textcolor{white}{.}\;\;\;\;\;\;\;\;\;\;\;\;\;\;\;\;\;\;\; \exists j \neq i\; ||H_{\theta^t}(A_{y=j}(\mathbf{x})) - \mathbf{\mu}_j||_2 \leq \tau_j \}$}
        \vspace{0.3em}
        \STATE{$A_F(B) \gets \{A_{y=j}(\mathbf{x})|\; (\mathbf{x}, i) \in B,\linebreak
        \textcolor{white}{.}\;\;\;\;\;\;\;\;\;\;\;\;\;\;\;\;\;\;\; \forall j \neq i\; ||H_{\theta^t}(A_{y=j}(\mathbf{x})) - \mathbf{\mu}_j||_2 > \tau_j \}$}
        \vspace{0.25em}
        \STATE{\textcolor{white}{.}\tcp{Phase 2. \circled{2}, \bcircled{2} of \autoref{fig-method-training-test}}}
        \STATE{$B^t_S \gets \{(\mathbf{x}, k)|\mathbf{x} \in A_S(B) \} $}
        \vspace{0.3em}
        \STATE{$B^t_F \gets \{(A_{y=j}(\mathbf{x}), i)|\; (\mathbf{x}, i) \in \mathcal{D}, \newline
        \textcolor{white}{.}\;\;\;\;\;\;\;\;\;\;\;\;\;\;\;\;\;\;\; \exists j \neq i \;  A_{y=j}(\mathbf{x}) \in A_F(\mathcal{D})
\} $} 
        \STATE{$B^t \gets B_{S}^t \cup B_{F}^t \cup B $} 
        \STATE{\textbf{TRAIN} $H_{\theta^t}$ to minimize $\mathcal{L}_{\text{fix}}(B^t;\theta^t)$}
        \STATE{$t \gets t + 1$}
    \ENDWHILE
     
\end{algorithmic}
  \caption{Training process of our adversarial detector.}
    \label{training-algorithm}
\end{algorithm}

    


\section{Related Work}
There is a broad range of literature in adversarial and robust learning, 
depending on objectives of attacks and defenses, or threat models~\cite{survey_2}.
Here we introduce and compare literature closely related to adversarial detection, 
and robust prediction.
\subsection{Adversarial detectors}
Adversarial detectors aim to detect adversarial examples before 
they compromise main tasks such as classifications.
Depending on base mechanisms of the detectors, 
we can categorize them into three types.
\paragraph{Auxiliary discriminative detectors.}
Early researchers introduced auxiliary \textit{discriminative} detector model $p(y=\textit{adversarial}|x)$.
It can be implemented as an additional class of adversarial examples in existing discriminative classifier~\cite{grosse-detect,dustbin},
 or  binary classification model~\cite{gong-detect, metzen2017detecting}.
Then the detector is trained to discriminate adversarial examples from benign examples.
However they are shown not effective against adaptive whitebox attacks 
where adversaries know everything about defenses and models~\cite{easily}.
We analyze this defect is from a limitation of discriminative approaches
where they do not consider likelihood $p(\mathbf{x},y)$ (how likely $(\mathbf{x}, y)$ exists).
Even though $p(\mathbf{x},y)$ is very low, the discriminative approaches
get to give high classification probability $p(y_j|\mathbf{x})=\frac{p(\mathbf{x},y_j)}{\sum_{i}p(\mathbf{x},y_i)}$,
if $p(\mathbf{x},y_j)$ of a class is \textit{relatively} higher than $p(\mathbf{x},y_i)$ of other classes.
On the other hand, our approach directly estimates \textit{likelihoods} of representations
$p(\mathbf{z}\Equal H_\theta(\mathbf{x}), y_i)$
and does not suffer from the problem against adaptive whitebox attacks.

\paragraph{Likelihood or density based detectors.} Recent research introduced defenses based on generative models ~\cite{simple,aregenerative,resisting,towardmnist,rethinking,capsnet} or
density estimation techniques~\cite{kd-detect, surprise, lid} which aim to 
detect adversarial examples based on likelihoods.
One type of them leverages distance metrics such as Mahalanobis distance
in a representation space and detects adversarial examples 
if a distance between  $\mathbf{x}$ and a cluster of benign data is larger than
a threshold.
However, as long as the overlaps of representations exist (as we found in motivational section \ref{motivation_section}),
the distance metric can not be an effective method to detect adversarial examples.
Its vulnerability is also studied by a prior work~\cite{rethinking_attack} showing that 
the Mahalanobis distance based detector~\cite{rethinking} is still vulnerable in adaptive whitebox attack model.
The other type of them~\cite{resisting,capsnet} leverages reconstruction loss 
of an input from encoder-decoder structures.
It detects an input if the reconstruction loss of the input
is larger than a predefined threshold.
However reconstruction loss is not effective as input data
become complex such as CIFAR10 and ImageNet. 
The vulnerability of this type of detector is studied in a prior~\cite{congennotroubst}.
In this paper, while the above approaches~\cite{simple, rethinking} only focus on clusters of benign representations,
we discover overlapping clusters of adversarial representations which
reveals fundamental limitations of the distance based detectors.
To resolve the problem we propose a new detector reducing the overlaps effectively detecting 
adversarial examples against adaptive whitebox attacks.
\paragraph{Statistical characteristics based detectors.} Besides the above approaches,
many detectors are proposed suggesting new statistical characteristics of
adversarial examples which can differentiate them from
the benign examples~\cite{oddodd, magnet, nic, feature-squeeze, filter-statistics}.
For example a detector~\cite{feature-squeeze} computes 
classification probabilities of an input before and after
applying input transformations.
If the difference between the two probabilities is large, it regards the input as adversarial.
The other detector~\cite{nic} analyzes activation patterns (representations) of neurons
of a classifier and compares the patterns to find different differences between benign and
adversarial examples.
However most of them are bypassed by adaptive whitebox attackers ~\cite{easily,ensemble-bypass,are-odd}.
Considering they rely on distance metrics on the representation space, we explain these failures 
are from the limitation of overlapping clusters of adversarial representations 
(section \ref{motivation_section}).
In this paper we resolve this problem by reducing the overlaps between benign and adversarial representations.

\subsection{Robust prediction}
\label{relwk}
In contrast to the detectors, robust prediction methods 
try to classify adversarial examples to their true labels.
We categorize them into three types according to their
base approaches.
\paragraph{Adversarial training.} Adversarial training
approaches~\cite{fgsm,pgd,trade} generate adversarial examples
with attacks, such as FGSM and PGD,
and augment them in training data.
The augmented adversarial examples are re-labeled to
their original labels so that they are
correctly classified after the adversarial training.
The adversarial training is empirically known as an effective 
robust prediction method which works against adaptive whitebox adversaries.
They show robust classification accuracy regardless of types of attacks,
if norms of perturbations $\delta$ are 
covered in the training process.
On the other hand, the performances of adversarial training method 
significantly degrade if the perturbation $\delta$ is not covered.
For example, a prior work \cite{elasticnet} found 
adversarial training methods can be broken by 
unseen perturbations $\delta$ although $\ell_1$ of $\delta$ is small.
On the other hand, our approach shows stable performances over 
various $\ell_p$ norm perturbations which implies
the effectiveness of cluster separation (\autoref{whitebox-experiment}).

\paragraph{Certified robust inferences.}
To achieve guaranteed robustness against any attacks 
rather than empirical evidence,
recent certified methods for robust inference rigorously
prove there is no adversarial example with $\ell_p$ norm 
perturbations given inputs $\mathbf{x}$.
They leverage several certification methods
including satisfiability modulo theories~\cite{smt-carlini, smt-ehlers, smt-hwang, smt-katz},
mixed integer programming~\cite{mit-bunel,mit-cheng,mit-dutta,mit-fischetti,mit-lumo}
, interval bound propagation~\cite{ibp,crown-ibp}
, constraining Lipschitz constant of neural networks~\cite{lipschitz-anil, lipschitz-cisse, lipschitz-gouk,lipschitz-tsuzuku} and randomized smoothing ~\cite{rs-cohen,rs-differential,rs-macer,rs-adv}.
However those guarantees cost a lot of training or testing time for the verification,
so many of them can not scale for practical applications of neural networks.
Also some of certified methods~\cite{smt-katz, lipschitz-cisse} degrade performances
of neural networks constraining their robustness properties.
Instead of taking the drawbacks of certifications,
we propose a practical method for adversarial detection
based on the cluster separation of adversarial representations.
\paragraph{Statistical defenses.} As like the statistical detectors, 
many robust prediction methods suggested statistical properties of adversarial examples~\cite{sap,mitigating_randomness,countering,defensegan,pixeldefend2018,defensegan}.
For instance, a defense method~\cite{sap} learns  
which neurons of a model to be deactivated
within a random process and improves the 
robustness of the model against a simple FGSM attack.
Another defense method~\cite{mitigating_randomness} also leverages
randomness in input pre-processing including random re-sizing and
random padding.
But later they are turned out to be ineffective,
just obfuscating gradients to disturb generations of adversarial examples.
Attack methods such as backward pass differentiable approximation (BPDA)
and expectation over transformations (EoT) are developed to 
break the obfuscated gradients~\cite{obfuscated}.
In contrast, our approach does not rely on randomness or
obfuscated gradients.
We assure this by showing blackbox 
attacks do not outperform adaptive whitebox attacks (\autoref{paragraph-blackbox})
as suggested in the paper ~\cite{obfuscated}.
\subsection{Analysis of adversarial examples}
Some of previous works studied root causes of adversarial attacks.
It is suggested that recent neural networks are very linear and
the linearity makes the networks very sensitive to small perturbations~\cite{fgsm}.
Meanwhile, assuming types of features, a dichotomy of robust feature
and non-robust feature is proposed~\cite{odd}. 
They analyze the intrinsic trade-off between robustness and accuracy.
Interestingly in a following work~\cite{nrf}, it is found that non-robust features suffice for achieving good accuracy on benign examples. 
In contrast the work defines vulnerable features as mathematical functions,
we define adversarial examples as clusters of representations.
Our definition enables us to assume separable
clusters of adversarial representation,
and further provide a robust detector effective against
adaptive whitebox attacks.

\section{Experiment}
\begin{table*}[t]
    \centering
        \caption{Summary of baselines including our detector.
        For LID, MAHA, Madry and TRADES, we access codes provided by authors and use their pre-trained models
        for the evaluations. Others are re-implemented based on descriptions on papers. 
        Resisting we used Convolutional VAE with 10 layers (MNIST) and 12 layers (CIFAR10).
        We use the same models provided by TRADES\cite{trade} for other approaches including ours.
        }
    \resizebox{1.0\textwidth}{!}{
    \setlength\tabcolsep{1.0pt}
\begin{tabular}{|l|l|l|c|c|c|c|c|}\hline
&\textbf{Detection metric / Detect if} &\textbf{Adaptive objective} &\textbf{Orignal source?} &\textbf{Benign clusters} &\textbf{Adversarial clusters} &\textbf{MNIST model} &\textbf{CIFAR10 model} \\\hline\hline
\textbf{Ours} &(A) Distance to the closest representation cluster / High &\autoref{eq-adaptive-attack}, $q(\mathbf{x}) = ||H_{\theta}(\mathbf{x}) - \mu_{\hat{y}}||_2$ &- &\Checkmark &\textcolor{applegreen}{\CheckmarkBold}  &Small CNN &Wide Resnet \\\hline
\textbf{L-Ben} (ablation) &(A) / High &\autoref{eq-adaptive-attack}, $q(\mathbf{x}) = ||H_{\theta}(\mathbf{x}) - \mu_{\hat{y}}||_2$ &- &\Checkmark &- &Small CNN &Wide Resnet \\\hline
\textbf{Vanilla} &(B) Classification probability / Low &\autoref{eq-pgd-attack}, Logit classification loss~\cite{cw} &- &- &- &Small CNN &Wide Resnet \\\hline
\textbf{Resisting}~\cite{resisting} &(A) + Reconstruction error / High &\autoref{eq-adaptive-attack}, $q(\mathbf{x}) = ||H_{\theta}(\mathbf{x}) - \mu_{\hat{y}}||_2 + || \mathbf{x} - \textrm{rec}(\mathbf{x})||_2 $ &Re-implemented &\Checkmark &- &CNN\_VAE (10L) &CNN\_VAE(12L) \\\hline
\textbf{FS}~\cite{feature-squeeze} &Classification difference after input processing /High &\autoref{eq-adaptive-attack}, $q(\mathbf{x}) = ||H_\theta(\mathbf{x}) - H_\theta(\mathbf{x'})||_1 $ &Re-implemented & - & - &Small CNN &Wide Resnet \\\hline
\textbf{LID}~\cite{lid} &Local intrinsic dimensionality score / Low &\autoref{eq-adaptive-attack}, $q(\mathbf{x}) = $ LID-score($\mathbf{x}$) &Yes &\Checkmark &- &CNN (5L) &CNN (12L) \\\hline
\textbf{Madry}~\cite{pgd} &(B) / Low &\autoref{eq-pgd-attack}, Logit classification loss~\cite{cw}  &Yes &- &- &CNN  &w32-10 wide  \\\hline
\textbf{TRADES}~\cite{trade} &(B) / Low &\autoref{eq-pgd-attack}, Logit classification loss~\cite{cw} &Yes &- &- &Small CNN &Wide Resnet \\ \hline
\textbf{MAHA}~\cite{simple} &(A), but Mahalanobis distance / High &\autoref{eq-adaptive-attack}, $q(\mathbf{x}) = $ Maha-score($\mathbf{x}$) &Yes &\Checkmark &- &- &Resnet ~\cite{simple} \\
\hline
\end{tabular}
    }
\vspace*{-0.3cm}
    \label{table-baselines}
\end{table*}

We conduct robust evaluations of our detector against both of adaptive
whitebox attacks and blackbox attacks. 
With 4 perturbation types: $\ell_p, p\in\{\infty, 2, 1, 0\}$ and 
7 types of attacks: PGD~\cite{pgd}, MIM~\cite{mim}, CW~\cite{cw}, FGSM~\cite{fgsm}, Transfer~\cite{delving}, NES~\cite{ql}, Boundary~\cite{boundary}, 
we empirically validate effectiveness of the reduced overlaps between adversarial 
and benign clusters of representations for adversarial detection.
The attacks are implemented based on attacks
 publicly available from github repositories, Cleverhans \footnote{https://github.com/tensorflow/cleverhans} and Foolbox \footnote{https://github.com/bethgelab/foolbox}.
We implemented our detector model using PyTorch 1.6~\cite{pytorch} machine learning library.
We mainly conducted our experiments on a server equipped with 4 of GTX 1080 Ti GPUs, 
an Intel i9-7900X processor and 64GB of RAM.

\paragraph{Baselines.} To demonstrate effectiveness of our detector,
we compare our detector to 7 baselines summarized in \autoref{table-baselines}.
\textbf{Ours} is our detector trained to reduce overlaps between adversarial 
and benign clusters of representations.
\textbf{L-Ben} is for an ablation study.
It is an adversarial detector trained by re-labeling adversarial examples 
to the original labels (only conduct \bcircled{1} of \autoref{fig-method-training-test}).
\textbf{Resisting}~\cite{resisting} is a detector based on VAE~\cite{vae}.
It leverages a property that adversarial examples tend to have high reconstruction errors.
It also leverages distances to the closest clusters as a detection metric.
\textbf{FS} is a detector~\cite{feature-squeeze} which computes
differences in classification probabilities before and after some input processing.
We implemented the detector with median filters.
\textbf{LID} and \textbf{MAHA} are detectors based on 
different distance metrics such as Local Intrinsic
Dimensionality or Mahalanobis distances.
These baselines show the limitation from the overlapping clusters of representations.
\textbf{Vanilla} is a general discriminative classifier trained with cross entropy loss.
Although vanilla is not applied any defense, 
we can leverage classification probabilities of Vanilla to detect adversarial examples~\cite{ood}.
Vanilla shows a lower bound of detection performances.
\textbf{Madry} and \textbf{TRADES} are adversarial training methods which are not
proposed as a detection method.
However, they can be reformulated as detectors as Vanilla
and we include them for comprehensive experiments between state-of-the-arts.
Note that existing approaches do not deal with adversarial clusters.

\paragraph{Datasets.} We evaluate our detector models on 
two datasets, MNIST~\cite{mnist} and CIFAR10~\cite{cifar10}.
MNIST contains grayscale hand-written digits from 0 to 9.
This dataset is composed of a training set of 60,000 images and,
a test set of 10,000 images with input dimension of 28$\times$28.
CIFAR10~\cite{cifar10} contains 10 classes of real color images.
This dataset is composed of a training set of 50,000 images,
and a test set of 10,000 images  with input dimension of 32$\times$32$\times$3.
The values of data elements are scaled from 0.0 to 1.0.
We selected the datasets since the datasets
are representative benchmarks with which the baselines are tested.

\paragraph{Evaluation metric.} Performances of the baselines 
are computed in two evaluation metrics, Attack Success Ratio (\textbf{ASR}) 
and Expected ROC AUC scores (\textbf{EROC}). 
\textbf{ASR} evaluates practical defense performances of the baselines with a fixed detection threshold, 
while \textbf{EROC} reflects overall characteristics of detection metrics of
the baselines.

\textbf{ASR} ranges from 0 to 1 and a lower value of ASR implies better detection performance.
It is defined in the threat model section \ref{threatmodel}
and , in short, it is a success ratio of an attack
over test inputs which are classified as true classes 
and not falsely detected as adversarial before the attack.
Note that ASR can be different depending on 
how conservative a detection threshold is.
For instance, if a detection threshold is chosen 
to have 50\% of false positive ratio (50\% of benign examples are 
falsely detected) it is much harder to bypass the detector and results
in lower ASR value than when the threshold is chosen to have 1\% of false positives.
To reflect the above scenarios, we use suffix ``-p'' with ASR
so that ASR-p means ASR when a detection threshold is 
chosen with p\% of false positives.
\begin{figure}
    \centering
    \includegraphics[width=7.5cm]{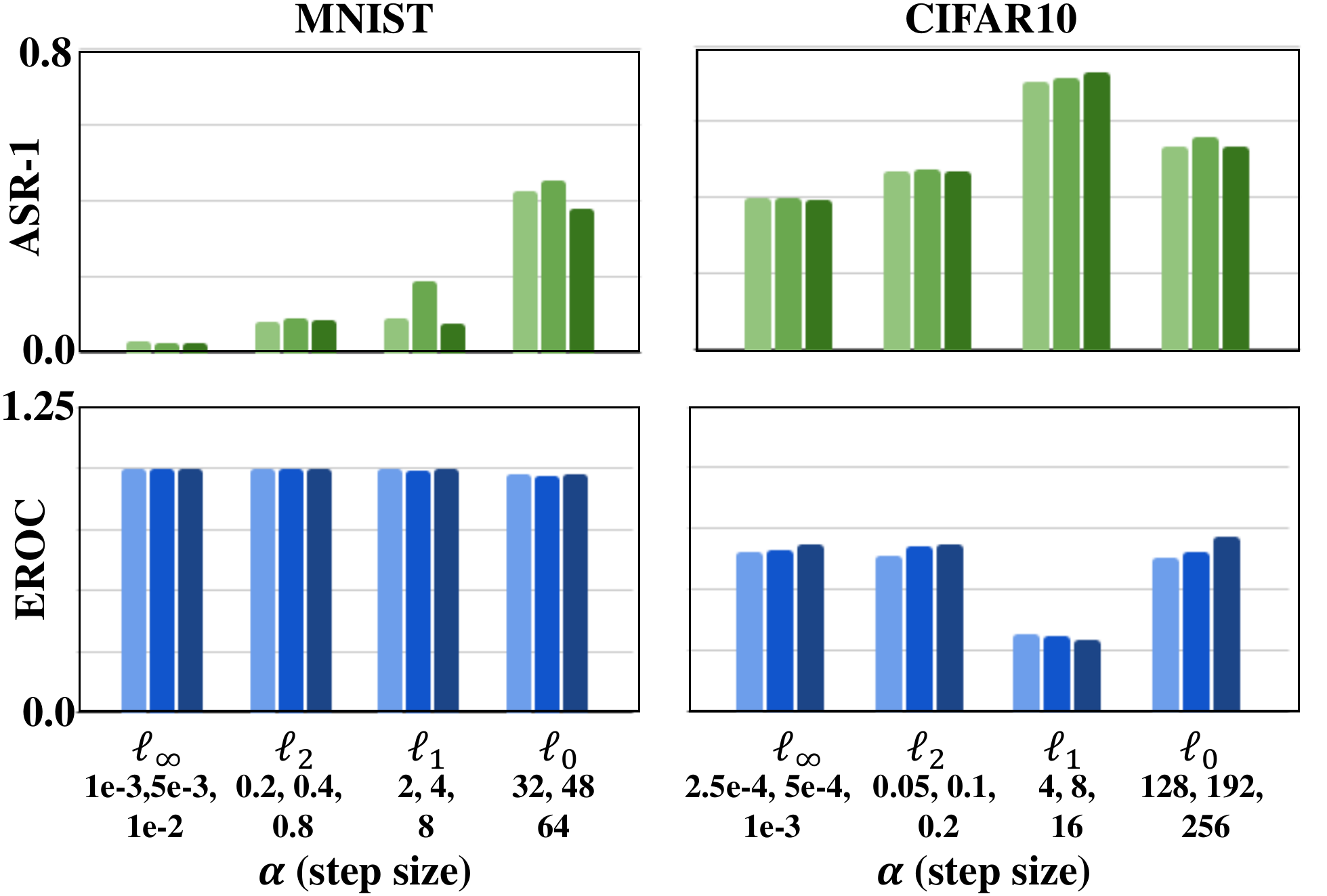}
\vspace{0.0em}
    \caption{Robustness of Ours to attack parameter $\alpha$ (step size). 
    Ours shows even or slight changes of detection performances.
    We select worst case $\alpha$ for the following experiments.}
    \label{fig-step-size}
\end{figure}

\textbf{EROC} ranges from 0 to 1 and a higher value of \textbf{EROC}
implies better detection performances.
EROC is proposed in this work as an extension of ROC AUC score~\cite{rocauc}.
ROC AUC score is a metric to evaluate
how clearly a detector distinguishes positive samples and negative samples.
However the metric is not compatible with our approach,
because it only deals with detectors with a single detection metric.
For example, our approach deal with one of $k$, the number of classes,
detection metrics depending on the closest benign cluster
given a representation (\autoref{paragraph-detection}).
So we have $k$ ROC AUC scores, one for each class.
To yield a single representative value, we get an expectation
of the $k$ ROC AUC scores by a weighted summation of them.
After an attack generates adversarial examples, 
a weight of ROC AUC score for class $i$ is computed as a fraction
of the number of adversarial examples classified to
class $i$ over the whole generated adversarial examples.

\begin{figure}
    \centering
    \includegraphics[width=7.5cm]{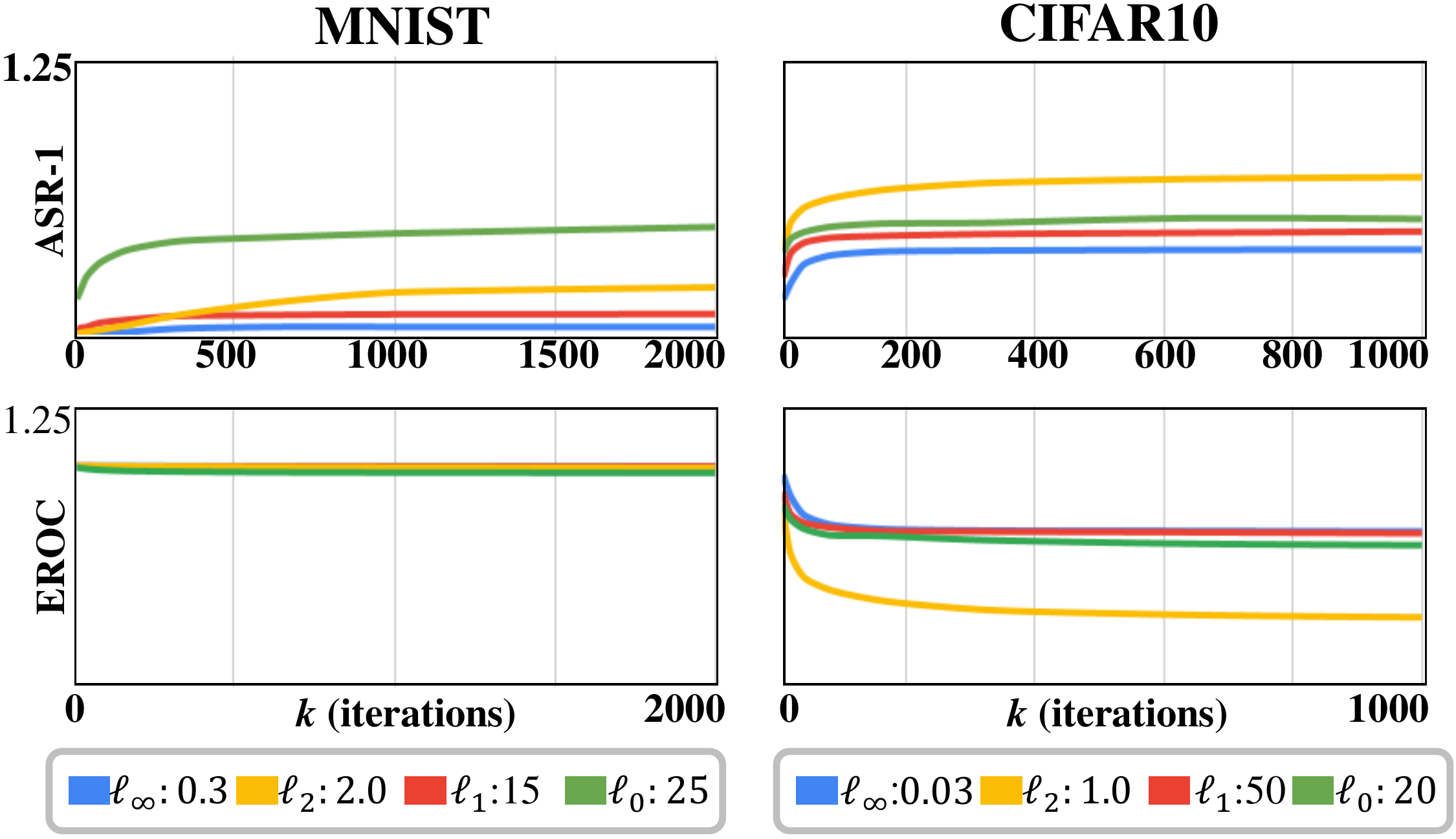}
\vspace{0.0em}
    \caption{Robustness of Ours to attack parameter $k$ (iterations).
    The detection performances converge around $k=1000$ for all types of perturbation
    and we choose $k=1000$ for attacks in the remaining experiments. }
    \label{fig-iters}
\end{figure}

\subsection{Adaptive Whitebox Attack Test}
\paragraph{Attack adaptation.} We evaluate the baselines against adaptive attacks,
and the attacks are implemented based on PGD attack~\cite{pgd}.
We summarize the adaptation methods for each baseline in the
``Adaptive objective'' column of \autoref{table-baselines}.
Basically adversaries adapt PGD attack by replacing
$q(\mathbf{x})$ of \autoref{eq-adaptive-attack} with 
differentiable detection metric of a baseline.
Depending on the detection conditions (``Detect if''), the sign of the
$q(\mathbf{x})$ can be negated.
For instance, an adversary can adapt PGD attack to Ours 
by replacing $q(\mathbf{x})$ of \autoref{eq-adaptive-attack} with Euclidean
distance between $H_\theta(\mathbf{x})$ and
to $\mathbf{\mu}_i$ of a wrong class $i$.
For the case of Vanilla, Madry and TRADES, we used Logit classification loss~\cite{cw}
to deal with vanishing gradient problems from soft-max normalization layer of neural networks.

\begin{table*}[t]
    \centering
        \caption{Adaptive whitebox attacks test.
            Robustness of the baselines are evaluated against adaptive PGD attacks,
            and adversarial examples are generated with 4 types of perturbation.
            We can validate the effect of cluster separation of 
            adversarial representations from the superior
            performances of Ours compared to L-Ben, Resisting, LID and MAHA
            which only consider clusters of benign representations.
            Although Madry and TRADES shows compatible performances on MNIST except PGD
            $\ell_\infty \Equal 0.4$, Ours outperforms them on CIFAR10 in regarding ASR-1.}
    \resizebox{1.0\textwidth}{!}{
    \setlength\tabcolsep{2.0pt}
\begin{tabular}{|c|c|cccccccc||c|c|ccccccccc|}
\multicolumn{10}{c}{\large{\textbf{MNIST}}} &\multicolumn{11}{c}{\large{\textbf{CIFAR10}}} \\\hline
& &\textbf{Ours} &\textbf{L-Ben} &\textbf{Vanilla} &\textbf{Resisting} &\textbf{FS} &\textbf{LID} &\textbf{Madry} &\textbf{TRADES} & & &\textbf{Ours} &\textbf{L-Ben} &\textbf{Vanilla} &\textbf{Resisting} &\textbf{FS} &\textbf{LID} &\textbf{Madry} &\textbf{TRADES} &\textbf{MAHA} \\\hline
&ACC. &0.97 &0.97 &\textbf{0.99} &\textbf{0.99} &0.97 &0.98 &0.98 &\textbf{0.99} & &ACC. &0.84 &0.87 &\textbf{0.98} &0.61 &0.87 &0.83 &0.87 &0.83 &0.93 \\ \hline \hline
\multirow{2}{*}{\shortstack{PGD\\$\ell_\infty \Equal 0.3$, T}} &ASR-1 $(\downarrow\downarrow)$ &\textbf{0.00} &0.01 &0.86 &0.72 &0.97 &0.93 &0.01 &\textbf{0.00} &\multirow{2}{*}{\shortstack{PGD\\$\ell_\infty \Equal 0.03$, T}} &ASR-1 $(\downarrow\downarrow)$ &0.19 &0.28 &1.00 &0.98 &1.00 &0.96 &0.15 &\textbf{0.11} &0.98 \\
&EROC$(\uparrow\uparrow)$ &\textbf{1.00} &\textbf{1.00} &0.98 &0.95 &0.36 &0.62 &\textbf{1.00} &\textbf{1.00} & &EROC$(\uparrow\uparrow)$ &0.91 &0.82 &0.45 &0.00 &0.03 &0.47 &\textbf{0.98} &\textbf{0.98} &0.06 \\ \hline
\multirow{2}{*}{\shortstack{PGD\\$ \ell_\infty \Equal 0.3$, U}} &ASR-1 $(\downarrow\downarrow)$ &\textbf{0.01} &0.06 &0.86 &0.89 &1.00 &0.98 &0.03 &\textbf{0.01} &\multirow{2}{*}{\shortstack{PGD\\$\ell_\infty \Equal 0.03$, U}} &ASR-1 $(\downarrow\downarrow)$ &\textbf{0.40} &0.55 &1.00 &0.99 &1.00 &1.00 &0.44 &\textbf{0.40} &1.00 \\
&EROC$(\uparrow\uparrow)$ &\textbf{1.00} &0.99 &0.98 &0.94 &0.09 &0.58 &\textbf{1.00} &\textbf{1.00} & &EROC$(\uparrow\uparrow)$ &0.69 &0.56 &0.47 &0.00 &0.01 &0.40 &0.89 &\textbf{0.91} &0.00 \\ \hline
\multirow{2}{*}{\shortstack{PGD\\$\ell_\infty \Equal 0.4$, T}} &ASR-1 $(\downarrow\downarrow)$ &\textbf{0.21} &0.50 &0.90 &0.85 &1.00 &0.93 &0.67 &0.73 &\multirow{2}{*}{\shortstack{PGD\\$\ell_\infty \Equal 0.06$, T}} &ASR-1 $(\downarrow\downarrow)$ &\textbf{0.32} &0.62 &1.00 &0.99 &1.00 &0.98 &0.57 &0.49 &0.95 \\
&EROC$(\uparrow\uparrow)$ &\textbf{0.99} &0.97 &0.98 &0.93 &0.10 &0.63 &0.87 &0.89 & &EROC$(\uparrow\uparrow)$ &0.76 &0.46 &0.45 &0.00 &0.00 &0.46 &0.79 &\textbf{0.88} &0.37 \\ \hline
\multirow{2}{*}{\shortstack{PGD\\$\ell_\infty \Equal 0.4$, U}} &ASR-1 $(\downarrow\downarrow)$ &\textbf{0.51} &0.85 &0.85 &0.95 &1.00 &0.98 &0.94 &0.95 &\multirow{2}{*}{\shortstack{PGD\\$\ell_\infty \Equal 0.06$, U}} &ASR-1 $(\downarrow\downarrow)$ &\textbf{0.61} &0.89 &1.00 &0.99 &1.00 &1.00 &0.80 &0.82 &1.00 \\
&EROC$(\uparrow\uparrow)$ &\textbf{0.98} &0.89 &\textbf{0.98} &0.90 &0.01 &0.57 &0.79 &0.88 & &EROC$(\uparrow\uparrow)$ &0.42 &0.16 &0.47 &0.00 &0.00 &0.00 &0.64 &\textbf{0.68} &0.00 \\ \hline
\multirow{2}{*}{\shortstack{PGD\\$\ell_2 \Equal 2.0$, T}} &ASR-1 $(\downarrow\downarrow)$ &0.03 &0.06 &0.39 &0.08 &0.51 &0.96 &0.03 &\textbf{0.01} &\multirow{2}{*}{\shortstack{PGD\\$\ell_2 \Equal 1.0$, T}} &ASR-1 $(\downarrow\downarrow)$ &\textbf{0.24} &0.35 &1.00 &1.00 &1.00 &1.00 &0.35 &0.33 &0.99 \\
&EROC$(\uparrow\uparrow)$ &\textbf{1.00} &0.99 &0.98 &\textbf{1.00} &0.89 &0.59 &\textbf{1.00} &\textbf{1.00} & &EROC$(\uparrow\uparrow)$ &0.89 &0.78 &0.40 &0.02 &0.01 &0.47 &0.91 &\textbf{0.94} &0.02 \\ \hline
\multirow{2}{*}{\shortstack{PGD\\$\ell_2 \Equal 2.0$, U}} &ASR-1 $(\downarrow\downarrow)$ &0.09 &0.21 &0.75 &0.27 &0.88 &1.00 &0.14 &\textbf{0.07} &\multirow{2}{*}{\shortstack{PGD\\$\ell_2 \Equal 1.0$, U}} &ASR-1 $(\downarrow\downarrow)$ &\textbf{0.48} &0.60 &1.00 &1.00 &1.00 &1.00 &0.67 &0.74 &1.00 \\
&EROC$(\uparrow\uparrow)$ &\textbf{1.00} &0.97 &0.91 &0.99 &0.70 &0.53 &0.99 &0.99 & &EROC$(\uparrow\uparrow)$ &0.68 &0.51 &0.43 &0.02 &0.01 &0.37 &\textbf{0.75} &\textbf{0.75} &0.00 \\ \hline
\multirow{2}{*}{\shortstack{PGD\\$\ell_1 \Equal 15$, T}} &ASR-1 $(\downarrow\downarrow)$ &0.04 &0.05 &0.36 &0.11 &0.46 &0.87 &\textbf{0.03} &0.07 &\multirow{2}{*}{\shortstack{PGD\\$\ell_1 \Equal 50$, T}} &ASR-1 $(\downarrow\downarrow)$ &\textbf{0.48} &0.70 &1.00 &1.00 &1.00 &1.00 &0.93 &0.92 &0.99 \\
&EROC$(\uparrow\uparrow)$ &\textbf{1.00} &0.99 &0.99 &\textbf{1.00} &0.91 &0.61 &\textbf{1.00} &\textbf{1.00} & &EROC$(\uparrow\uparrow)$ &\textbf{0.67} &0.42 &0.31 &0.00 &0.00 &0.45 &0.37 &0.47 &0.02 \\ \hline
\multirow{2}{*}{\shortstack{PGD\\$\ell_1 \Equal 15$, U}} &ASR-1 $(\downarrow\downarrow)$ &0.19 &0.11 &0.73 &0.37 &0.84 &0.98 &0.11 &\textbf{0.09} &\multirow{2}{*}{\shortstack{PGD\\$\ell_1 \Equal 50$, U}} &ASR-1 $(\downarrow\downarrow)$ &\textbf{0.72} &0.94 &1.00 &1.00 &1.00 &1.00 &0.98 &0.99 &1.00 \\
&EROC$(\uparrow\uparrow)$ &\textbf{0.99} &\textbf{0.99} &0.97 &0.98 &0.72 &0.56 &\textbf{0.99} &\textbf{0.99} & &EROC$(\uparrow\uparrow)$ &0.32 &0.12 &0.41 &0.00 &0.00 &0.37 &\textbf{0.47} &0.40 &0.00 \\ \hline
\multirow{2}{*}{\shortstack{PGD\\$\ell_0 \Equal 25$, T}} &ASR-1 $(\downarrow\downarrow)$ &0.17 &\textbf{0.03} &0.61 &0.16 &0.64 &0.81 &0.08 &0.04 &\multirow{2}{*}{\shortstack{PGD\\$\ell_0 \Equal 20$, T}} &ASR-1 $(\downarrow\downarrow)$ &0.29 &0.40 &0.83 &0.94 &0.29 &0.72 &\textbf{0.23} &0.28 &0.66 \\
&EROC$(\uparrow\uparrow)$ &0.99 &0.99 &0.93 &0.99 &0.85 &0.65 &0.99 &\textbf{1.00} & &EROC$(\uparrow\uparrow)$ &0.88 &0.82 &0.71 &0.00 &0.95 &0.52 &\textbf{0.96} &0.95 &0.78 \\ \hline
\multirow{2}{*}{\shortstack{PGD\\$\ell_0 \Equal 25$, U}} &ASR-1 $(\downarrow\downarrow)$ &0.45 &\textbf{0.12} &0.85 &0.68 &0.93 &0.98 &0.23 &0.19 &\multirow{2}{*}{\shortstack{PGD\\$\ell_0 \Equal 20$, U}} &ASR-1 $(\downarrow\downarrow)$ &0.53 &0.60 &0.98 &0.98 &\textbf{0.52} &0.92 &0.53 &0.64 &0.76 \\
&EROC$(\uparrow\uparrow)$ &0.97 &0.98 &0.82 &0.97 &0.64 &0.54 &0.97 &\textbf{0.99} & &EROC$(\uparrow\uparrow)$ &0.63 &0.57 &0.56 &0.00 &\textbf{0.89} &0.44 &\textbf{0.89} &0.79 &0.64 \\
\hline\hline
\multirow{2}{*}{\shortstack{Worst\\case}} &ASR-1 $(\downarrow\downarrow)$ &\textbf{0.51} &0.85 &0.90 &0.95 &1.00 &1.00 &0.94 &0.95 &\multirow{2}{*}{\shortstack{Worst\\case}} &ASR-1 $(\downarrow\downarrow)$ &\textbf{0.72} &0.94 &1.00 &1.00 &1.00 &1.00 &0.98 &0.99 &1.00 \\
&EROC ($\uparrow\uparrow$) &\textbf{0.97} &0.89 &0.82 &0.90 &0.01 &0.53 &0.79 &0.88 & &EROC ($\uparrow\uparrow$) &0.32 &0.12 &0.31 &0.00 &0.00 &0.00 &0.37 &\textbf{0.40} &0.00 \\\hline
\end{tabular}
    }
\vspace*{-0.3cm}
    \label{table-sota-adaptive-whitebox-attacks}
\end{table*}
\paragraph{Sensitiveness to attack parameters.} 
To demonstrate Ours is not sensitive to attack parameters,
and to select strong attack parameters to evaluate robustness,
we conduct two experiments.
Specifically we test attack parameters $\alpha$ and $k$ in \autoref{eq-pgd-attack}.
In the first experiment we test sensitiveness on $\alpha$
which is also called step size (\autoref{fig-step-size}).
We attacked Ours with the adaptive PGD attack varying $\alpha$ (x-axis).
We could check Ours shows even or slight performance changes on both of the 
datasets (left and right) and metrics (top and bottom). 
In the following experiments, we conduct adaptive attacks with the several number of $\alpha$ (3 to 8)
for all of the baselines and use the worst performances.
In the second experiment, we analyze the strength of the attack parameter $k$, iterations.
We increase $k$ from 0 to 2000 (MNIST) and from 0 to 1000 (CIFAR10),
and plot the result in \autoref{fig-iters}.
The ASR-1 and EROC for the both of the dataset converge before $k=1000$ (MNIST) and
$k=800$ (CIFAR10), so we choose $k=1000$ as an attack iteration in the following experiments.

\paragraph{Robustness to various types of perturbation.} 
\label{whitebox-experiment}
In \autoref{table-sota-adaptive-whitebox-attacks}, we evaluated 
the baselines under adaptive whitebox PGD attacks with
various perturbation types $\ell_{p}, p \in\{0, 1, 2, \infty\}$
for thorough comparisons of robustness between them.
The left half of the table is for MNIST and the
remaining half is for CIFAR10. 
On each half, the first column denotes types and sizes of perturbations, and 
whether an attack is targeted (T) or untargeted (U).
For example, ``PGD $\ell_\infty \Equal 0.3$, T'' means
the baselines are evaluated with adaptive PGD attack under 
perturbations $\delta$ whose $\ell_\infty$ norm are less than 0.3.
For each of the attacks, we measure ASR-1 and EROC metrics.
The down arrows of ASR-1 mean lower is better and
the up arrows of EROC means higher is better.
We stress the best performances with boldface.
Note that we measure conservative ASR-1 which means 
detection thresholds with 1\% of false positives.

\begin{table*}[t]
    \centering
        \caption{Blackbox attack test. Robustness of the baselines are evaluated against
        3 different blackbox attacks with $\ell_\infty$ and $\ell_2$ perturbations.
        Most of the baselines are effective in the blackbox attack model. }
    \resizebox{1.0\textwidth}{!}{
    \setlength\tabcolsep{2.0pt}
\begin{tabular}{|c|c|cccccccc||c|c|ccccccccc|}
\multicolumn{10}{c}{\large{\textbf{MNIST}}} &\multicolumn{11}{c}{\large{\textbf{CIFAR10}}}\\\hline
& &\textbf{Ours} &\textbf{L-Ben} &\textbf{Vanilla} &\textbf{Resisting} &\textbf{FS} &\textbf{LID} &\textbf{Madry} &\textbf{TRADES} &\textbf{} &\textbf{} &\textbf{Ours} &\textbf{L-Ben} &\textbf{Vanilla} &\textbf{Resisting} &\textbf{FS} &\textbf{LID} &\textbf{Madry} &\textbf{TRADES} &\textbf{MAHA} \\\hline \hline
\multirow{2}{*}{\shortstack{Transfer \\ $\ell_\infty \Equal 0.3$}} &ASR-1 $(\downarrow\downarrow)$ &\textbf{0.00} &\textbf{0.00} &0.01 &0.08 &0.96 &0.84 &\textbf{0.00} &\textbf{0.00} &\multirow{2}{*}{\shortstack{Transfer \\ $\ell_\infty \Equal 0.03$}} &ASR-1 $(\downarrow\downarrow)$ &0.02 &\textbf{0.00} &0.71 &0.15 &0.55 &0.52 &0.01 &0.01 &0.78 \\
&EROC$(\uparrow\uparrow)$ &\textbf{1.00} &0.99 &\textbf{1.00} &\textbf{1.00} &0.73 &0.82 &\textbf{1.00} &\textbf{1.00} & &EROC$(\uparrow\uparrow)$ &\textbf{1.00} &\textbf{1.00} &0.77 &0.97 &0.92 &0.78 &1.00 &0.99 &0.72 \\ \hline
\multirow{2}{*}{\shortstack {NES\\ $\ell_\infty \Equal 0.3$}} &ASR-1 $(\downarrow\downarrow)$ &\textbf{0.01} &0.05 &0.69 &0.54 &0.75 &0.92 &0.03 &\textbf{0.01} &\multirow{2}{*}{\shortstack {NES\\ $\ell_\infty \Equal 0.06$}} &ASR-1 $(\downarrow\downarrow)$ &\textbf{0.43} &0.55 &1.00 &0.99 &0.98 &0.99 &0.47 &0.46 &0.93 \\ 
&EROC$(\uparrow\uparrow)$ &\textbf{1.00} &0.99 &0.76 &0.98 &0.66 &0.69 &\textbf{1.00} &\textbf{1.00} & &EROC$(\uparrow\uparrow)$ &0.71 &0.59 &0.02 &0.00 &0.35 &0.69 &\textbf{0.88} &\textbf{0.88} &0.11 \\ \hline
\multirow{2}{*}{\shortstack{NES\\ $\ell_2 \Equal 2.0$}} &ASR-1 $(\downarrow\downarrow)$ &\textbf{0.01} &0.11 &0.60 &0.14 &0.44 &0.93 &0.05 &0.02 &\multirow{2}{*}{\shortstack{NES\\ $\ell_2 \Equal 1.0$}} &ASR-1 $(\downarrow\downarrow)$ &\textbf{0.37} &0.40 &1.00 &0.99 &0.95 &0.98 &0.55 &0.57 &0.65 \\
&EROC$(\uparrow\uparrow)$ &\textbf{1.00} &0.99 &0.84 &\textbf{1.00} &0.88 &0.59 &0.99 &\textbf{1.00} & &EROC$(\uparrow\uparrow)$ &0.77 &0.68 &0.02 &0.00 &0.35 &0.76 &\textbf{0.84} &0.83 &0.34 \\ \hline
\multirow{2}{*}{\shortstack{Boundary\\ $\ell_\infty \Equal 0.3$}} &ASR-1 $(\downarrow\downarrow)$ &\textbf{0.00} &\textbf{0.00} &\textbf{0.00} &\textbf{0.00} &0.54 &0.89 &\textbf{0.00} &\textbf{0.00} &\multirow{2}{*}{\shortstack{Boundary\\ $\ell_\infty \Equal 0.06$}} &ASR-1 $(\downarrow\downarrow)$ &\textbf{0.00} &\textbf{0.00} &0.02 &0.34 &0.40 &0.17 &\textbf{0.00} &0.04 &0.34 \\
&EROC$(\uparrow\uparrow)$ &\textbf{1.00} &\textbf{1.00} &\textbf{1.00} &\textbf{1.00} &0.96 &0.75 &\textbf{1.00} &\textbf{1.00} & &EROC$(\uparrow\uparrow)$ &\textbf{1.00} &\textbf{1.00} &\textbf{1.00} &0.95 &0.93 &0.90 &\textbf{1.00} &0.98 &0.92 \\ \hline
\multirow{2}{*}{\shortstack{Boundary\\ $\ell_2 \Equal 2.0$}} &ASR-1 $(\downarrow\downarrow)$ &\textbf{0.00} &\textbf{0.00} &\textbf{0.00} &\textbf{0.00} &0.49 &0.82 &\textbf{0.00} &\textbf{0.00} &\multirow{2}{*}{\shortstack{Boundary\\ $\ell_2 \Equal 1.0$}} &ASR-1 $(\downarrow\downarrow)$ &0.02 &0.00 &0.01 &0.14 &0.43 &0.06 &\textbf{0.00} &0.05 & 0.38 \\
&EROC$(\uparrow\uparrow)$ &\textbf{1.00} &\textbf{1.00} &\textbf{1.00} &\textbf{1.00} &0.96 &0.77 &\textbf{1.00} &\textbf{1.00} & &EROC$(\uparrow\uparrow)$ &\textbf{1.00} &0.99 &\textbf{1.00} &0.98 &0.93 &0.95 &\textbf{1.00} &0.98 &0.92 \\
\hline
\end{tabular}

    }
\vspace*{-0.3cm}
    \label{table-sota-adaptive-blackbox-attacks}
\end{table*}

\begin{table}[t]
    \centering
        \caption{Adaptive whitebox attack tests with different detection thresholds and attack methods.}
    \resizebox{1.0\columnwidth}{!}{
    \setlength\tabcolsep{1.5pt}
\begin{tabular}{|c|c|cccc||c|c|cccc|}
\multicolumn{6}{c}{\textbf{MNIST}} &\multicolumn{6}{c}{\textbf{CIFAR10}} \\\hline
& &\textbf{Ours} &\textbf{L-Ben} &\textbf{Vanilla} &\textbf{TRADES} &\textbf{} &\textbf{} &\textbf{Ours} &\textbf{L-Ben} &\textbf{Vanilla} &\textbf{TRADES} \\ \hline\hline
\multirow{3}{*}{\shortstack{PGD \\ $\ell_\infty \Equal 0.4$} } &ASR-2 &\textbf{0.32} &0.69 &0.56 &0.89 &\multirow{3}{*}{\shortstack{PGD \\ $\ell_\infty \Equal 0.06$} } &ASR-5 &\textbf{0.58} &0.87 &1.00 &0.79 \\
&ASR-5 &0.09 &0.48 &\textbf{0.01} &0.59 & &ASR-10 &\textbf{0.56} &0.86 &1.00 &0.72 \\
&EROC &\textbf{0.98} &0.88 &\textbf{0.98} &0.87 & &EROC &0.42 &0.16 &0.47 &\textbf{0.68} \\ \hline
\multirow{3}{*}{\shortstack{PGD \\ $\ell_2 \Equal 2.0$} } &ASR-2 &0.05 &0.17 &0.58 &\textbf{0.04} &\multirow{3}{*}{\shortstack{PGD \\ $\ell_2 \Equal 1.0$} } &ASR-5 &\textbf{0.41} &0.56 &1.00 &0.70 \\
&ASR-5 &\textbf{0.02} &0.09 &0.19 &\textbf{0.02} & &ASR-10 &\textbf{0.34} &0.52 &1.00 &0.63 \\
&EROC &\textbf{1.00} &0.97 &0.91 &0.99 & &EROC &0.68 &0.51 &0.43 &\textbf{0.75} \\ \hline
\multirow{3}{*}{\shortstack{PGD \\ $\ell_0 \Equal 25$} } &ASR-2 &0.36 &\textbf{0.10} &0.68 &0.14 &\multirow{3}{*}{\shortstack{PGD \\ $\ell_0 \Equal 20$} } &ASR-5 &\textbf{0.53} &0.60 &0.96 &0.58 \\
&ASR-5 &0.19 &\textbf{0.04} &0.26 &0.08 & &ASR-10 &\textbf{0.48} &0.52 &0.92 &0.49 \\ 
&EROC &0.97 &\textbf{0.98} &0.82 &\textbf{0.99} & &EROC &0.63 &0.57 &0.56 &\textbf{0.79} \\ \hline \hline
\multirow{3}{*}{\shortstack{CW \\ $\ell_2 \Equal 2.0$} } &ASR-2 &0.57 &0.17 &0.56 &\textbf{0.07} &\multirow{3}{*}{\shortstack{CW \\ $\ell_2 \Equal 1.0$} } &ASR-2 &\textbf{0.46} &0.69 &1.00 &0.59 \\
&ASR-5 &0.43 &0.11 &0.53 &0.03 & &ASR-5 &\textbf{0.37} &0.66 &1.00 &0.54 \\
&EROC &0.93 &0.97 &0.78 &\textbf{0.99} & &EROC &0.80 &0.58 &0.20 &\textbf{0.88} \\ \hline
\multirow{3}{*}{\shortstack{MIM \\ $\ell_\infty \Equal 0.4$} } &ASR-2 &\textbf{0.49} &0.62 &0.76 &0.71 &\multirow{3}{*}{\shortstack{MIM \\ $\ell_\infty \Equal 0.06$} } &ASR-5 &\textbf{0.68} &0.90 &0.96 &0.81 \\
&ASR-5 &\textbf{0.23} &0.42 &0.41 &0.47 & &ASR-10 &\textbf{0.66} &0.89 &0.93 &0.79 \\
&EROC &\textbf{0.96} &0.91 &0.72 &0.92 & &EROC &0.36 &0.12 &0.47 &\textbf{0.68} \\ \hline
\multirow{3}{*}{\shortstack{FGSM \\ $\ell_\infty \Equal 0.4$} } &ASR-2 &\textbf{0.00} &0.01 &0.21 &0.01 &\multirow{3}{*}{\shortstack{FGSM \\ $\ell_\infty \Equal 0.06$} } &ASR-5 &\textbf{0.00} &\textbf{0.00} &0.58 &0.43 \\
&ASR-5 &\textbf{0.00} &\textbf{0.00} &0.13 &\textbf{0.00} & &ASR-10 &\textbf{0.00} &\textbf{0.00} &0.40 &0.35 \\
&EROC &\textbf{1.00} &\textbf{1.00} &0.94 &\textbf{1.00} & &EROC &\textbf{0.99} &\textbf{0.99} &0.92 &0.92 \\ \hline \hline
\multirow{3}{*}{\shortstack{PGD \\ $\ell_\infty \Equal 1.0$} } &ASR-2 &1.00 &1.00 &0.56 &0.95 &\multirow{3}{*}{\shortstack{PGD \\ $\ell_\infty \Equal 1.0$} } &ASR-5 &1.00 &1.00 &1.00 &1.00 \\
&ASR-5 &0.99 &1.00 &0.01 &0.81 & &ASR-10 &1.00 &1.00 &0.99 &1.00 \\
&EROC &0.40 &0.12 &0.98 &0.53 & &EROC &0.00 &0.00 &0.39 &0.26 \\ \hline
\multirow{3}{*}{\shortstack{PGD \\ $\ell_2 \Equal 6.0$} } &ASR-2 &0.94 &0.90 &0.66 &0.89 &\multirow{3}{*}{\shortstack{PGD \\ $\ell_2 \Equal 6.0$} } &ASR-5 &1.00 &0.97 &0.99 &1.00 \\
&ASR-5 &0.85 &0.75 &0.26 &0.72 & &ASR-10 &1.00 &0.95 &0.98 &1.00 \\
&EROC &0.63 &0.51 &0.76 &0.74 & &EROC &0.01 &0.25 &0.46 &0.26 \\
\hline
\end{tabular}
    }
\vspace*{-0.3cm}
    \label{table-sota-other-attacks}
\end{table}
In the case of MNIST, Ours performs well across  
many types of perturbations with low ASR-1 and high EROC.
Most of the ASR-1 are not higher than 0.20 and 
the lowest EROC is 0.97.
Although Madry and TRADES are proposed as robust inference 
methods~(\autoref{relwk}), they achieve 
good detection performances across the perturbation types.
We regard it is because they are trained to 
lower classification probabilities (detection metric) of adversarial examples in their 
training process.
However the performances of the robust inference methods
show steep degradation up to ASR of 0.95 and EROC 0.79 
if they are attacked with unseen size of perturbations (PGD $\ell_\infty \Equal 0.4$, U).
In contrast, Ours shows less performance reduction (ASR-1 of 0.51 and EROC of 0.98).
Compared to the L-Ben, which is for an ablation study to check
effectiveness of separating clusters of adversarial representations,
Ours performs superior to L-Ben for the all perturbation types
except PGD $\ell_0\Equal 25$ and PGD $\ell_1\Equal 25$, U. 
This result validates the cluster separation between
adversarial representations and benign representations
reduces the overlaps between them and improves detection
performances.
As we observed in the motivational section (\autoref{motivation_section}),
the detection method based on distances between representations 
(Resisting, LID, MAHA) does not show effective performances
under adaptive whitebox attacks. 
In the case of CIFAR10, we observe consistent results
with MNIST dataset. The robust inference methods (Madry, TRADES)
achieved good detection metrics across the perturbation types
while showing large degradation in the case of unseen size of perturbations
(PGD, $\ell_\infty \Equal 0.06$, U).
The ablation study between Ours and L-ben still holds in CIFAR10, 
even improving performances on the perturbation types $\ell_0$ and $\ell_1$.

In summary, Ours achieves improves worst case ASR-1 by 45\% (MNIST)
and 27\% (CIFAR10) over the existing adversarial detectors and
robust inference methods.
We also validate our argument about separation of adversarial clusters
in a representation space is effective with the comparisons to L-ben
and three representation-distance based detectors (Resisting, LID, MAHA).

\paragraph{Robustness to thresholds and attack methods.} 
We analyze the detectors more closely 
with different detection thresholds and against different attack methods.
We previously analyzed the performances of the baseline detectors
with very low 1\% false positives thresholds 
to conservatively compare them.
However the conservative analysis has a limitation
that it can not evaluate how well a detection approach
reduces ASR as increasing the threshold.
So we compare the ASR reduction property varying
the thresholds in \autoref{table-sota-other-attacks}.
We only demonstrate a subset of the baselines for compact comparisons. 
First of all, we analyze the baselines under 
attacks where they show relatively high ASR-1 compared
to the other attacks in \autoref{table-sota-adaptive-whitebox-attacks}.
They correspond to the first three rows of attacks in
\autoref{table-sota-other-attacks}. 
Note that we conduct all attacks in untargeted way and
measured two ASR with different thresholds.
In case of MNIST, under the PGD $\ell_\infty \Equal 0.4$, Ours shows 0.09 of ASR-5 which is
much lower than 0.59 of TRADES.
Vanilla reduces to 0.01 of ASR while does not reduce ASR on CIFAR10.

Next, we evaluate the baselines against different attack methods.
The methods include Carlini \& Wager (CW)~\cite{easily},
Momentum Iterative Method (MIM)~\cite{mim} and Fast Gradient Sign Method (FGSM)~\cite{fgsm}
and we adapt them to each baseline.
The second three rows of \autoref{table-sota-other-attacks} 
show the results of the attacks.
We conduct CW attack finding the best confidence parameters with 5 binary searches.
For the MIM attack we use momentum decay factor with 0.9.
We find CW attack is more powerful than 
PGD attack to Ours, showing higher ASR on MNIST.
However Ours still retains EROC of 0.93 which shows
its effectiveness.
On the other hand, Ours achieves the best ASR-2 and ASR-5 score
against CW attacks on CIFAR10. 
Against MIM $\ell_{\infty}$, we can check consistent results 
with PGD $\ell_{\infty}$ showing a little performance degradation
from all baselines.
Under FGSM $\ell_{\infty}$, which performs a single step perturbation,
all baselines perform well except TRADES shows High ASR-5
and low EROC.

In addition to the above evaluations, we attack the baselines with
large perturbations $\ell_\infty\Equal 1.0$ and $\ell_2\Equal 6.0$
for validating Ours does not depend on obfuscated gradients 
, as suggested in a prior work ~\cite{obfuscated}.
We checked ASRs of Ours reach near 1.00 under the large
perturbation attacks which confirm obfuscated gradient
problem does not occur in case of Ours.

\subsection{Blackbox Attack Test}
\label{paragraph-blackbox}
We evaluate the defenses under three blackbox attacks
in \autoref{table-sota-adaptive-blackbox-attacks}
and the details of the attacks are explained in 
\autoref{paragraph-blackbox-attacks}.
This experiment provides another evidence that Ours does not 
exploit obfuscated gradients~\cite{obfuscated}.
Overall, most of the baselines performs well against the 
blackbox attacks while FS and LID are not effective.
In the case of FS, median filter alone is not enough 
detection mechanism~\cite{feature-squeeze,ensemble-bypass} even though
the attack is not adapted to FS (Transfer).
LID is suggested to be used with an auxiliary discriminative detector,
however the detection score alone can not discriminate 
adversarial examples from benign examples.
We validate there is no obfuscated gradients problem,
verifying no blackbox attacks perform better than
adaptive whitebox attacks.

\section{Conclusion}
In this work, we introduced a new adversarial detector under 
the novel hypothesis of separable clusters 
of adversarial representations 
and we proposed to \emph{learn} the separable
clusters formulating an optimization problem 
of deep neural networks.
Specifically, we modeled the representation space as
a mixture of Gaussians.
We validated our model under both of the adaptive whitebox attacks
and blackbox attacks.
Studying more effective assumptions on the distribution
of representations and applying our defense to different 
tasks would be interesting future research. 





\bibliographystyle{plain}
\bibliography{references}

\end{document}